\documentclass[11pt]{article}       
\usepackage{geometry}               
\usepackage{comment}
\usepackage{tikz}
\usetikzlibrary{shapes.geometric, arrows}
\usepackage{adjustbox}

\usepackage{graphicx}
\usepackage{caption}
\usepackage{subcaption}
\usepackage{comment}
\usepackage{amsmath} 
\usepackage{amssymb}  
\usepackage{amsthm}  
\usepackage{bm}
\usepackage{lipsum}
\usepackage{color}
\usepackage{multirow}
\usepackage{array}
\usepackage{cite}
\usepackage{hyperref}

\usepackage{algorithm}
\usepackage{algpseudocode}

\numberwithin{equation}{section}

\title{A diffusion-based generative model for financial time series via geometric Brownian motion}

\author{
Gihun Kim\thanks{Equal contribution. Department of Mathematics, POSTECH}
\and
Sun-Yong Choi\thanks{Equal contribution. Department of Financial Mathematics, Gachon University}
\and
Yeoneung Kim\thanks{Corresponding author. Department of Applied Artificial Intelligence, SeoulTech, yeoneung@seoultech.ac.kr}
\and
}

\begin{document}
\maketitle

\pagestyle{myheadings}
\thispagestyle{plain}

\begin{abstract}
We propose a novel diffusion-based generative framework for financial time series that incorporates geometric Brownian motion (GBM), the foundation of the Black--Scholes theory, into the forward noising process. Unlike standard score-based models that treat price trajectories as generic numerical sequences, our method injects noise proportionally to asset prices at each time step, reflecting the heteroskedasticity observed in financial time series. By accurately balancing the drift and diffusion terms, we show that the resulting log-price process reduces to a variance-exploding stochastic differential equation, aligning with the formulation in score-based generative models. The reverse-time generative process is trained via denoising score matching using a Transformer-based architecture adapted from the Conditional Score-based Diffusion Imputation (CSDI) framework. Empirical evaluations on historical stock data demonstrate that our model reproduces key stylized facts such as heavy-tailed return distributions, volatility clustering, and the leverage effect more realistically than conventional diffusion models.
\end{abstract}


\section{Introduction}
The generation of realistic synthetic financial time series has become a critical task in modern quantitative finance, with applications in risk modeling, trading strategy development, and stress testing. As markets become increasingly data-driven, the demand for generative models capable of producing plausible and high-fidelity trajectories has intensified. Deep generative models such as Generative Adversarial Networks (GANs)~\cite{Goo14} and Variational Autoencoders (VAEs)~\cite{kingma2013auto} have shown strong performance in generating high-dimensional data and have been applied to simulate asset price dynamics, capturing some statistical features like heavy tails and volatility clustering. However, most of these approaches treat financial time series as generic numerical sequences, often overlook domain-specific structures, leading to unrealistic outputs under distributional shifts or extreme market conditions.

Recently, diffusion-based generative models~\cite{ho2020denoising}, originally developed for image generation, have gained significant attention for modeling complex data distributions by defining both forward (noising) and reverse (denoising) processes as a Markov chain. Subsequently, score-based generative models~\cite{song2020score} were proposed as a continuous-time generalization of diffusion models, and have shown remarkable capability in generating high-dimensional financial sequences across a wide range of generation tasks. A central idea of diffusion models is to progressively corrupt data with Gaussian noise, mapping the complex data distribution to a simple prior, typically a standard Gaussian. Generation is then performed by reversing this process: the score function, i.e., the gradient of the log-density, is estimated at each time step via denoising score matching, allowing a sample from the prior to be gradually transformed into realistic data. While this framework is well-suited for general high-dimensional data, it does not leverage the domain-specific structure of financial time series. In particular, most existing methods adopt forward stochastic differential equations (SDEs) that assume additive Gaussian noise, an unrealistic assumption for asset prices, which evolve multiplicatively and exhibit volatility that scales with price level. This motivates the integration of geometric Brownian motion (GBM) from the Black--Scholes (BS) model for the forward noising process.

{  
The Black--Scholes model~\cite{black1973pricing} stands as a cornerstone in financial mathematics and engineering, and more broadly, in the financial market. Prior to the seminal work of Black and Scholes, option pricing relied heavily on intuition and ad-hoc methods, often resulting in market inefficiencies and exploitable arbitrage opportunities. The introduction of the BS model not only provided closed-form solutions for the theoretical pricing of European call and put options, but also established a quantitative framework for understanding and managing derivative instruments. Key metrics derived from the model-such as implied volatility, delta, and gamma—have become indispensable tools for market participants.

Despite certain limitations, such as the assumption of constant volatility and the model’s limited capacity to account for leverage effects, the BS model remains dominant due to its theoretical soundness (i.e., arbitrage-free pricing), mathematical tractability, and the standardization and communication benefits it brings to the market. These attributes have made the model essential for the smooth functioning of modern derivative markets.

Beyond option valuation, the applications of the BS model extend to various areas in finance. One notable example is delta hedging, where the model plays a central role in constructing dynamic hedging strategies. In corporate finance, it is widely employed to value employee stock options, warrants, and convertible securities. In portfolio management, the BS model offers a standardized means to assess expected returns and associated risks, enabling investors to optimize their portfolios in alignment with their risk preferences and return objectives. Additionally, the model’s underlying principles have been adapted to value real options in capital budgeting and strategic decision-making contexts.
}

In this work, we propose a novel generative framework that explicitly incorporates GBM, the stochastic process underpinning the BS model, into the forward noising process of a score-based model. By injecting Gaussian noise in log-price space, our method induces volatility that scales with asset price, which is a key source of heteroskedasticity observed in real-world markets. This heteroskedasticity can be viewed as arising from state-dependent volatility, a structural feature we aim to capture through our forward diffusion design. To this end, we balance the drift and diffusion terms such that the forward process reduces to a variance-exploding SDE in log-price space, which aligns with standard score-based generative modeling frameworks. For the reverse process, we adopt a Transformer-based architecture inspired by the Conditional Score-based Diffusion Imputation (CSDI) model~\cite{tashiro2021csdi}, enabling flexible modeling of temporal dependencies. Empirically, our GBM-inspired framework captures key stylized facts of financial time series such as heavy tails, volatility clustering, and leverage effects more effectively than non-GBM diffusion frameworks.

{ 
This approach represents a significant departure from conventional studies that assume a specific stochastic differential equation (SDE) process for stock prices and use it to simulate or forecast price dynamics. Whereas prior research typically aims to generate stylized features of stock prices by directly modeling the underlying price process, our study adopts a novel perspective by incorporating GBM-based noising into the price process itself. This integration of geometric Brownian motion into the forward noising mechanism constitutes a meaningful contribution to the modeling of financial time series.
}

\subsection{Related works}

{  
A growing number of recent studies have used GAN-based frameworks for financial time series forecasting. For instance, \cite{takahashi2019modeling} utilized a GAN-based framework to model financial time series data and successfully reproduced key stylized facts such as heavy-tailed distributions, volatility clustering, and leverage effects. Similarly, \cite{tovar2020deep} developed a GAN-based model to generate synthetic financial data that closely resembles real market behavior. The model was designed to capture both upward and downward market trends and to enhance the accuracy of stock price forecasting. In a related study, \cite{rizzato2023generative} proposed a conditional financial scenario generation model that integrates a Bidirectional Generative Adversarial Network (BiGAN) architecture with a Markov Chain Monte Carlo (MCMC) technique. Furthermore, \cite{li2024enhancing} proposed a GAN-driven stock prediction model, demonstrating that the synthetic data produced by GANs can effectively reflect market sentiment and volatility.

Traditional machine learning and time series forecasting methods often fall short in providing probabilistic forecasts and typically rely on strong distributional assumptions regarding future target variables, limiting their ability to effectively capture uncertainty~\cite{vuletic2024fin}. In contrast, the predictive results presented in the aforementioned studies demonstrate that the use of GAN-based models can overcome these limitations. Specifically, GANs are shown to effectively capture key stylized facts of financial markets and enable the simulation of realistic financial data, thereby offering a more flexible and robust approach to financial time series modeling.
}

In addition to these GAN-based approaches, a growing body of research has explored more general-purpose generative models for time series, including classical autoregressive models, RNNs, and VAEs. These models laid the foundation for recent advances in generative modeling, but often lacked the ability to reproduce the complex stochastic structures observed in financial markets.

{  
Several recent studies have introduced innovative applications of score-based diffusion models across various domains. For instance, \cite{tashiro2021csdi} proposed a novel time series imputation method, Conditional Score-based Diffusion for Imputation (CSDI), which incorporates conditioning on observed data within the diffusion process. Their empirical analysis, conducted on healthcare and environmental datasets, demonstrates that the proposed approach significantly outperforms existing methods in time series interpolation and probabilistic forecasting tasks.

\cite{richemond2022categorical} extended the score-based diffusion framework to categorical data by proposing a method that directly diffuses data points located on an $n$-dimensional probability simplex. Their approach utilizes the Cox--Ingersoll--Ross (CIR) process, a model commonly employed in mathematical finance, to enable diffusion within the simplex structure.

In a related study, \cite{ouyang2023missdiff} introduced MissDiff, a diffusion-based generative framework designed to learn directly from tabular data with missing values. Their method eliminates the need for imputation or deletion by handling missing data natively within the diffusion process. Numerical experiments validate the empirical effectiveness of MissDiff, highlighting its advantages over traditional missing data handling techniques.

}
While these models have achieved impressive sample quality, their lack of inductive bias rooted in financial theory often leads to the generation of implausible or unstable dynamics, particularly under distributional shifts or market stress. This limitation has motivated recent trends such as physics-informed generative modeling and neural SDEs, which aim to embed structural assumptions into the model architecture. Nevertheless, few works have explicitly grounded diffusion-based generative models in fundamental asset pricing theory, such as the BS framework. This work addresses the gap by embedding GBM, central to the Black--Scholes framework, into the generative process, thereby improving both practical relevance and theoretical clarity.

\subsection{Our contribution}

We make the following key contributions:

\begin{itemize}
    \item We propose a novel score-based generative framework that integrates geometric Brownian motion (GBM) into the forward noising process. By balancing the drift and diffusion terms, we show that the log-price dynamics reduce to a variance-exploding SDE compatible with standard diffusion modeling.

    \item We adapt a Transformer-based diffusion model by incorporating explicit temporal encodings and refining the architecture to better capture localized volatility structures and long-range temporal dependencies in financial returns.

    \item We investigate how different noise schedules (linear, exponential, cosine) interact with VE, VP, and GBM-based forward processes, revealing that GBM combined with exponential or cosine scheduling best reproduces characteristics of financial data.

    \item Our model consistently reproduces heavy-tailed return distributions, volatility clustering, and the leverage effect, demonstrating improved alignment with real-world financial data over existing methods such as GANs and baseline diffusion models.
\end{itemize}

\section{Preliminary}
\subsection{Diffusion probabilistic models}
The original diffusion probabilistic models (DPMs)~\cite{ho2020denoising} introduce a forward process that corrupts data over a finite time horizon with discrete time steps. The central idea is to model the data generation process as the reversal of a diffusion process that gradually perturbs the data with noise until it converges to a known prior distribution, typically a Gaussian. A neural network is then trained to reverse this process, or equivalently, to predict the added noise given corrupted data. Score-based generative models extend this framework by introducing a continuous-time diffusion process that incrementally adds noise to the data throughout the time horizon. In this section, we review the framework of score-based generative modeling to clarify its connection to the methodology proposed in this paper.

Let $ x_0 \sim p_{\text{data}} $ denote a sample from the data distribution, where $ x_0 \in \mathbb{R}^d $. The forward process defines a sequence of random variables $ \{x_t\}_{t \in [0,T]} \subset \mathbb{R}^d $ governed by the SDE
\[
    \mathrm{d}x_t = f(x_t,t) \mathrm{d}t + g(t) dW_t,
\]
where $ f(x_t,t)\in \mathbb{R}^d $ is the drift term, $g(t) \in \mathbb{R}$ is the diffusion coefficient, and $ W_t \in \mathbb{R}^d $ denotes a standard $ d $-dimensional Brownian motion. As the time step $t$ increases, the process gradually transforms the original data distribution $ p_{\text{data}} $ into a known prior distribution $ p_T $, which is usually set to be the standard Gaussian distribution, $ \mathcal{N}(0, I_d) $.

In contrast, the reverse process, which aims to generate realistic samples, is defined by evolving the SDE backward in time. According to Anderson's theorem~\cite{anderson1982reverse}, under mild regularity conditions on $f$ and $g$, the reverse-time SDE takes the form:
\[
    \mathrm{d}x_t = \left[f(x_t,t) - g(t)^2 \nabla_{x_t} \log p_t(x_t)\right] \mathrm{d}t + g(t) \mathrm{d}\overline{W}_t,
\]
where $ \overline{W}_t $ denotes a reversed-time Brownian motion, and $ \nabla_{x_t} \log p_t(x_t) $ is called the score function of the perturbed distribution at time $ t $.

Two commonly employed forward diffusion processes are the variance-preserving (VP) and variance-exploding (VE) SDEs. The VP SDE maintains the total variance of the process and is defined as
\begin{equation}
    \mathrm{d}x_t = -\frac{1}{2} \sigma_t^2 x_t\, \mathrm{d}t + \sigma_t \mathrm{d}W_t, \quad x_0 \sim p_{\text{data}},
\end{equation}
where $ \sigma_t > 0 $ is a time-dependent noise schedule and $ W_t $ is a standard Brownian motion. The marginal distribution at time $ t $ is given by
\begin{equation}
    x_t \overset{D}{=} \sqrt{\alpha_t} x_0 + \sqrt{1 - \alpha_t}\, \varepsilon_t, \quad
    \alpha_t := \exp\left(-\int_0^t \sigma_s^2\, \mathrm{d}s\right), \quad \varepsilon_t \sim \mathcal{N}(0, I).
\end{equation}

On the other hand, in the case of VE SDE, the total variance increases over time and is given by
\begin{equation}
    \mathrm{d}x_t = \sqrt{\frac{\mathrm{d}[\sigma_t^2]}{\mathrm{d}t}} \mathrm{d}W_t, \quad x_0 \sim p_{\text{data}},
\end{equation}
where $ \sigma_t^2 $ is an increasing variance function such that $ \sigma_0^2 = 0 $. Then, the marginal distribution becomes
\begin{equation}
    x_t \sim \mathcal{N}(x_0, \sigma_t^2 I).
\end{equation}

Both VP and VE SDEs define forward processes that can be reversed to define generative models. Given a score estimate $ s_\theta(x, t) \approx \nabla \log p_t(x) $, the corresponding reverse-time SDE for the VP formulation is:
\begin{equation}
    \mathrm{d}x_t = \sigma_t^2 \left( s_\theta(x_t, t) - \frac{1}{2} x_t \right) \mathrm{d}t + \sigma_t\, \mathrm{d}\overline{W}_t.
\end{equation}
For VE SDE, the reverse-time SDE becomes:
\[
\mathrm{d}x(t) = - \frac{d[\sigma_t^2]}{dt} s_\theta(x_t, t) \, \mathrm{d}t + \sqrt{ \frac{\mathrm{d}[\sigma_t^2]}{\mathrm{d}t} } \, \mathrm{d}\overline{W}_t.
\]
Here, $ \overline{W}_t$ denotes the reverse-time Brownian motion as above.

Since the score function \( \nabla_{x_t} \log p_t(x_t) \) is typically intractable, it is approximated by a neural network \( s_\theta(x, t) \), and is trained to minimize a score-matching objective. A common approach is denoising score matching (DSM), which in the continuous-time setting leads to the following loss:
\[
\mathbb{E}_{t \sim \mathcal{U}(0,T)} \mathbb{E}_{x_0 \sim p_{\text{data}},\, x_t \sim p_{t|0}} \left[ \lambda(t) \left\| s_\theta(x_t, t) - \nabla_{x_t} \log p_{t|0}(x_t \mid x_0) \right\|^2 \right],
\]
where \( \lambda(t) \) is a time-dependent weighting function and \( p_{t|0}(x_t \mid x_0) \) is the transition kernel of the forward SDE. For common processes such as VP and VE, the target score \( \nabla_{x_t} \log p_{t|0} \) is known analytically, allowing efficient training via stochastic gradient descent. Once trained, the score network enables sample generation by solving the corresponding reverse-time SDE, starting from \( x_T \sim p_T \) and integrating backward to \( x_0 \).

{
  
\subsection{Statistical properties financial time seires}
Financial time series exhibit a number of empirically observed regularities, often referred to as stylized facts, which any realistic generative model should aim to reproduce. These properties are universal across a wide range of financial assets and markets and have been documented in both econometric and econophysical literature. In this section, we review several key statistical characteristics of financial time series of our interest, which will be used as evaluation criteria for our proposed model.
}

\paragraph{(i) Heavy-tailed distributions.}
Empirical asset return distributions exhibit significant deviations from normality, especially in the tails. The return distribution often exhibits power-law decay, with probability density $ P(r) \sim |r|^{-\alpha} $, where the tail exponent $ \alpha $ typically lies in the range $ 3 \leq \alpha \leq 5 $. This property captures the frequent occurrence of extreme price movements.

\paragraph{(ii) Volatility clustering.}
Despite the absence of linear autocorrelation in raw returns, the magnitude of returns (or volatility) displays persistent temporal dependence. Periods of large returns tend to be followed by further large returns (regardless of sign), and similarly for small returns, resulting in long-range dependence in the absolute returns:
\[
\mathrm{Corr}(|r_t|, |r_{t+k}|) \sim k^{-\beta}, \quad 0.1 \leq \beta \leq 0.5.
\]

\paragraph{(iii) Leverage effect.}
In many equity markets, negative returns are often followed by periods of increased volatility, a phenomenon known as the leverage effect. It manifests as a negative correlation between past returns and future volatility, which can be measured via the lead–lag correlation:
\[
L(k) = \frac{\mathbb{E}[r_t |r_{t+k}|^2 - r_t |r_t|^2]}{\mathbb{E}[|r_t|^2]^2}.
\]

The aforementioned characteristics play a critical role in the derivatives market. First, the presence of heavy-tailed distributions must be explicitly accounted for in risk measurement to avoid underestimating the probability of extreme losses, particularly when using metrics such as Value-at-Risk (VaR) and Expected Shortfall (ES) \cite{harmantzis2006empirical}. Second, volatility clustering has led to the formulation of stochastic volatility models, which better accommodate the persistence of volatility over time \cite{jacquier2002bayesian}. Lastly, the leverage effect is closely associated with the implied volatility skew or smile observed in options markets, providing a theoretical basis for such phenomena \cite{dennis2006stock}. Collectively, these three stylized facts, heavy tails, volatility clustering, and the leverage effect, are essential considerations in both the theoretical modeling and practical application of derivatives pricing and risk management.

\section{Methodology}

We propose a score-based generative model for financial time series that injects Gaussian noise in log-price space, inspired by the structure of GBM. This design reflects key features of asset prices, including their multiplicative nature and the resulting heteroskedasticity. Formally, we apply the forward diffusion process to log-price vector \( X_0 = \{\log s_1, \dots, \log s_L\} \) to generate the perturbed time series \( X_t \). The exponential mapping back to price space induces state-dependent volatility, consistent with empirical observations in financial markets. To this end, we balance the drift and diffusion terms so that the forward process reduces to a variance-exploding SDE in log-price space. 

Instead of directly modeling the price process to reproduce stylized facts, we introduce GBM-based noising into the process, providing a principled framework for modeling key empirical characteristics of financial time series.
\subsection{Geometric Brownian motion in score-based Models}

\begin{figure}[ht]
    \centering
    \includegraphics[width=0.8\linewidth]{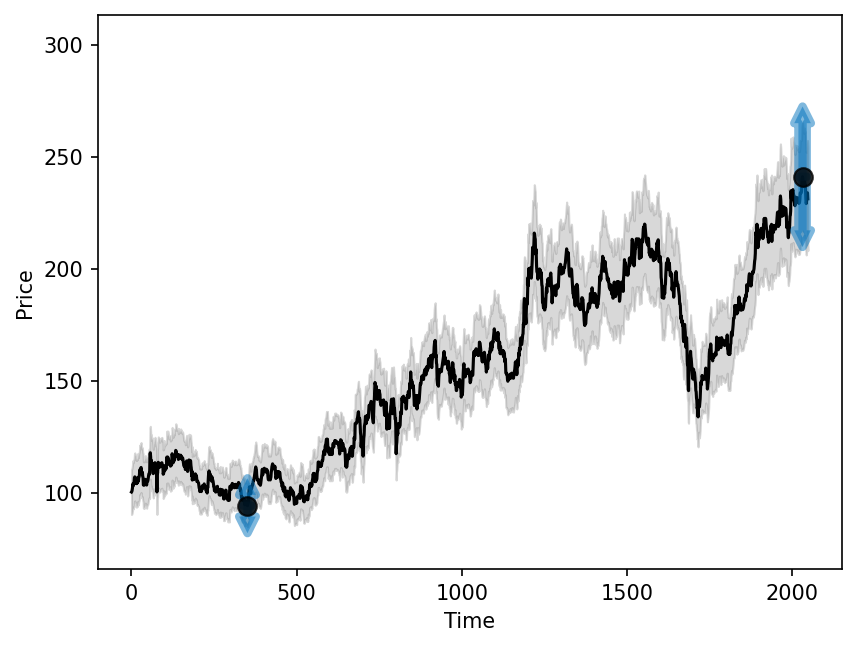}
    \caption{Generated price time series with a shaded envelope indicating price-dependent noise intensity. The width of the envelope reflects the level of uncertainty, which increases with the stock price.}
    \label{fig:arrow_timeseries}
\end{figure}

In the Black--Scholes framework, the asset price at time $\tau>0$ denoted by $s_\tau$ evolves according to a geometric Brownian motion:
\begin{equation}
\mathrm{d}s_\tau = \tilde \mu_\tau s_\tau  \mathrm{d}\tau + \tilde \sigma_\tau s_\tau  \mathrm{d}\tilde w_\tau,
\label{eq:bs-sde}
\end{equation}
where $\tilde w_\tau \in \mathbb{R}$ is a one-dimensional Brownian motion, $\tilde \sigma_t > 0$ denotes the volatility, and $\tilde \mu_\tau$ is the drift term. The multiplicative structure of the noise implies that volatility scales proportionally with the asset price.

To incorporate this into the framework of diffusion-based generative model, we discretize the time series $\{s_\tau\}_{\tau \geq 0}$ as a vector $S_{0} = \{s_{1:L}\} \in \mathbb{R}^{1 \times L}$ of length $L$, and apply GBM-inspired noise over time to obtain a perturbed series $S_t \in \mathbb{R}^{1\times L}$, which is given by 
\[
\mathrm{d} S_t =  \mu_t S_t \mathrm{d}\tau +  \sigma_t S_t \mathrm{d}  W_t,
\]
where $W_t$ denotes the $L$-dimensional Brownian motion. Transforming $S_t$ to log-prices $X_t = \log S_t$ and applying the Ito's lemma:
\begin{equation}
\begin{split}
\mathrm{d}X_t &= \frac{1}{S_t}\,\mathrm{d}S_t - \frac{1}{2S_t^2} (\mathrm{d}S_t)^2 \\
&= \left(  \mu_t - \tfrac{1}{2} \sigma_t^2 \right) \mathrm{d}t +  \sigma_t \mathrm{d}W_t,
\end{split}
\label{eq:log-sde}
\end{equation}
where we use $(\mathrm{d}S_t)^2 = \sigma_t^2 S_t^2 \, \mathrm{d}t$. Choosing $\mu_t = \tfrac{1}{2} \sigma_t^2$, the drift term is elimiated, yielding the VE SDE:
\begin{equation}
\mathrm{d}X_t = \sigma_t \, \mathrm{d}W_t.
\label{eq:ve-sde}
\end{equation}

We set $\sigma_t = \sqrt{\beta_t}$ with $\beta_0 = 0$, yielding the forward SDE:
\[
\mathrm{d}X_t = \sqrt{\beta_t} \, \mathrm{d}W_t, \quad X_t = \log S_t.
\]    

As illustrated in Figure~\ref{fig:arrow_timeseries}, the GBM-based formulation naturally injects larger noise into higher-value regimes, thereby preserving economically meaningful dynamics in the generated trajectories. It inherently scales noise with the underlying price level, enhancing the model's sensitivity to large market movements. As a result, the diffusion process can more accurately capture stylized facts such as volatility clustering and heavy tails.

The reverse-time generative process is trained via denoising score matching, identical to the VE diffusion model in~\cite{song2020score}, using a Transformer-based architecture adapted from the CSDI framework.

\subsubsection{Architecture of the score network}

\begin{figure}[ht]
    \centering
    \includegraphics[width=1\linewidth]{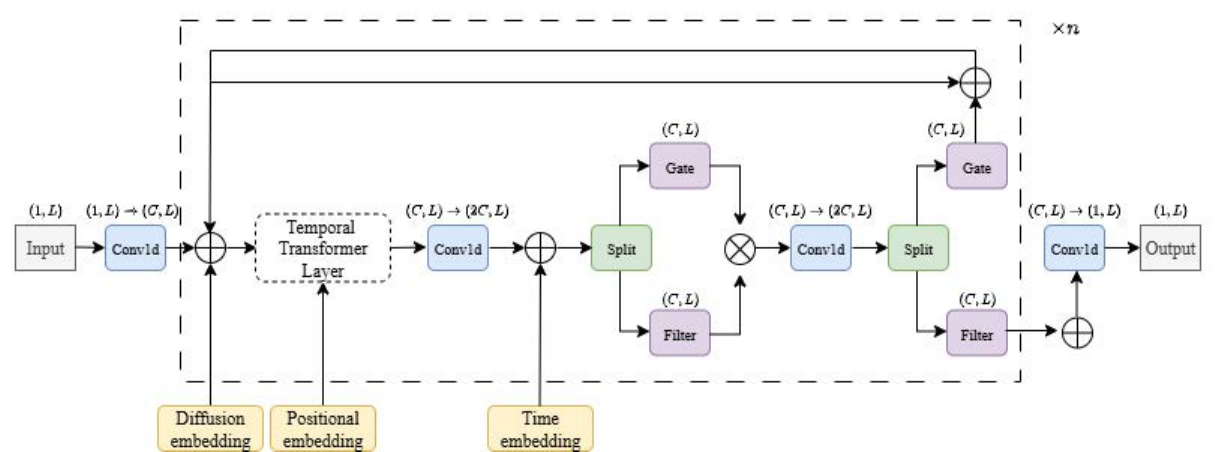}
    \caption{Neural network architecture based on CSDI used for score estimation. The network consists of convolutional layers, transformer blocks, gated residual modules, and skip connections to model financial time series data.}
    \label{fig:csdi_architecture}
\end{figure}

We adopt the neural network architecture originally developed in the Conditional Score-based Diffusion Model (CSDI) framework~\cite{tashiro2021csdi}, designed for conditional time series generation from partially observed data. The model combines convolutional encoders with self-attention mechanisms to capture both local structure and long-range dependencies. A 1D convolution layer first projects this input into a high-dimensional latent space. Temporal information is then incorporated using three types of embeddings. The diffusion step embedding encodes the noise level at each step, allowing the network to adjust its behavior across different corruption scales. The time embedding represents absolute temporal positions through a combination of sinusoidal functions and learnable vectors. The positional embedding encodes the relative position of each element within the input sequence, enabling the model to capture short-term sequential patterns.

These embeddings are added to the latent representation and passed through a Transformer block, which models global dependencies across time. The resulting features are further processed through a series of gated residual blocks ($n=4$)~\cite{kong2020diffwave}, which include convolutional layers, gating mechanisms, and skip connections to enable stable training and hierarchical feature extraction. The final score estimate \(s_\theta(x_t, t)\) is obtained by aggregating the outputs of all residual blocks via skip connections and projecting them back to the input dimension through a final 1D convolutional layer.

To adapt the architecture for our financial time series data, we introduce two main modifications: explicit positional encodings are applied prior to the Transformer layers, and model hyperparameters such as channel widths and hidden dimensions are fine-tuned to better capture stylized empirical features of asset returns. The vanilla CSDI implementation proposed in \cite{tashiro2021csdi} employs 64 convolution channels, a 128-dimensional diffusion‐step embedding, and a 16-dimensional feature embedding. With these settings, the model reproduces volatility clustering and heavy tails reasonably well, but fails to capture the leverage effect with high fidelity. We found that the insufficient expressiveness of the diffusion-step and feature embeddings impaired the model’s ability to encode the asymmetric covariance structure between negative returns and future volatility.

Consequently, we increased the representational capacity to 128 convolution channels, 256-dimensional diffusion embeddings, and 64-dimensional feature embeddings. The higher-dimensional diffusion embedding facilitates learning of steeper score gradients at early noise levels, and the enlarged feature embedding separates low-frequency drift from high-frequency leverage shocks. We empirically validate the impact of these architectural modifications, with detailed results presented in Section~\ref{subsec:effect}.

\subsubsection{Collection of Data}

To train and evaluate our generative model, we constructed a large-scale dataset of historical stock prices from the U.S. equity market. In particular, we utilized the constituents of the S\&P 500 index, which represent a diverse and liquid set of publicly traded companies \footnote{The data were collected using the \texttt{yfinance} API, which provides convenient access to historical price records.}.

We first obtained the list of S\&P 500 constituents and excluded tickers with problematic symbols (e.g., \texttt{BRK.B}, \texttt{BF.B}) that do not follow a standard format. For each remaining ticker, we downloaded the maximum available historical data at daily frequency. Since our goal is to capture long-term financial dynamics, we filtered the stocks to retain only those whose available price history extends more than 40 years back. This selection criterion ensures that the dataset encompasses diverse market regimes, including various boom and bust cycles, thus providing a rich context for training the generative model.

For each selected stock, we calculated daily log-returns, defined as $r_t = \log (p_t / p_{t-1})$, where $p_t$ represents the adjusted closing price on day $t$. The log-returns were computed after removing missing values to ensure consistency in the time series.

To create training samples suitable for diffusion modeling, we applied a sliding window approach to the log-return sequences. Each subsequence was extracted with a fixed length of 2048 time steps and a stride of 400 time steps, resulting in partially overlapping subsequences. Stocks with log-return histories shorter than the required subsequence length were discarded from the sample extraction process. This procedure yielded a large number of diverse subsequences covering various stocks and time periods.

\section{Experimental results}
\begin{figure}[t]
  \centering

  \begin{subfigure}[t]{0.32\textwidth}
    \includegraphics[width=\linewidth]{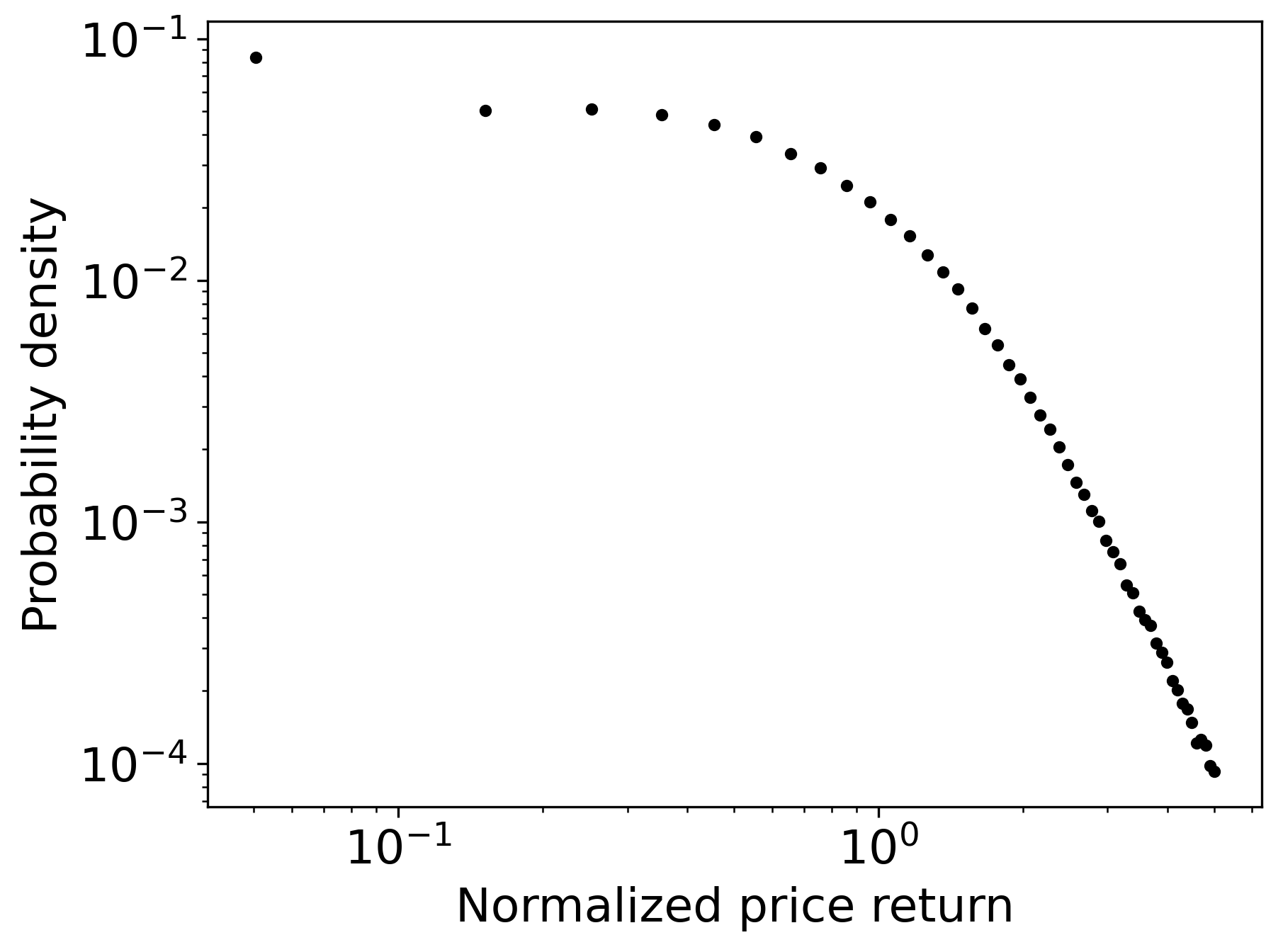}
    \caption{Heavy-tail distribution with $\alpha=4.35$}
    
  \end{subfigure}\hfill
  \begin{subfigure}[t]{0.32\textwidth}
    \includegraphics[width=\linewidth]{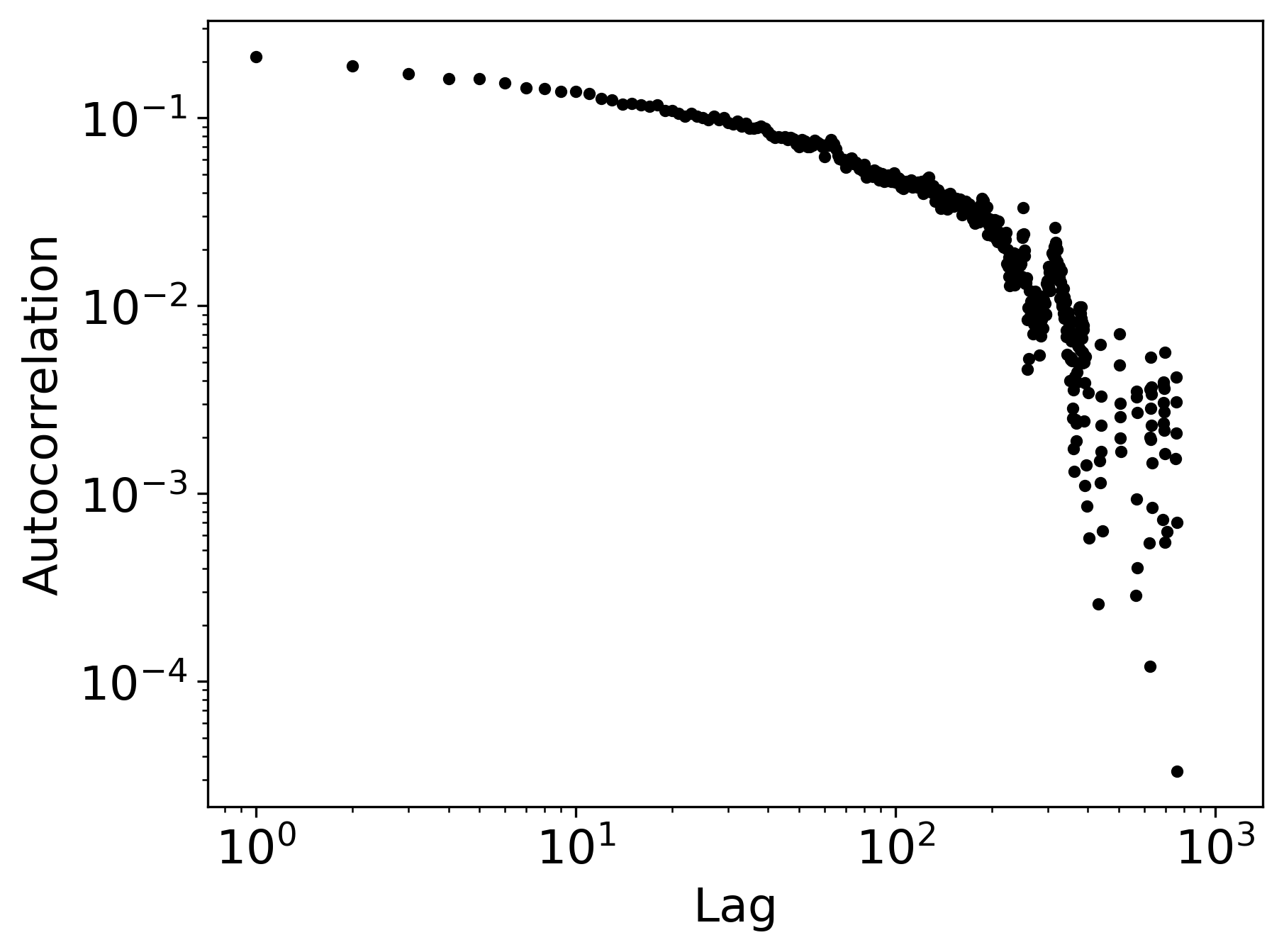}
    \caption{Volatility clustering}
    
  \end{subfigure}\hfill
  \begin{subfigure}[t]{0.32\textwidth}
    \includegraphics[width=\linewidth]{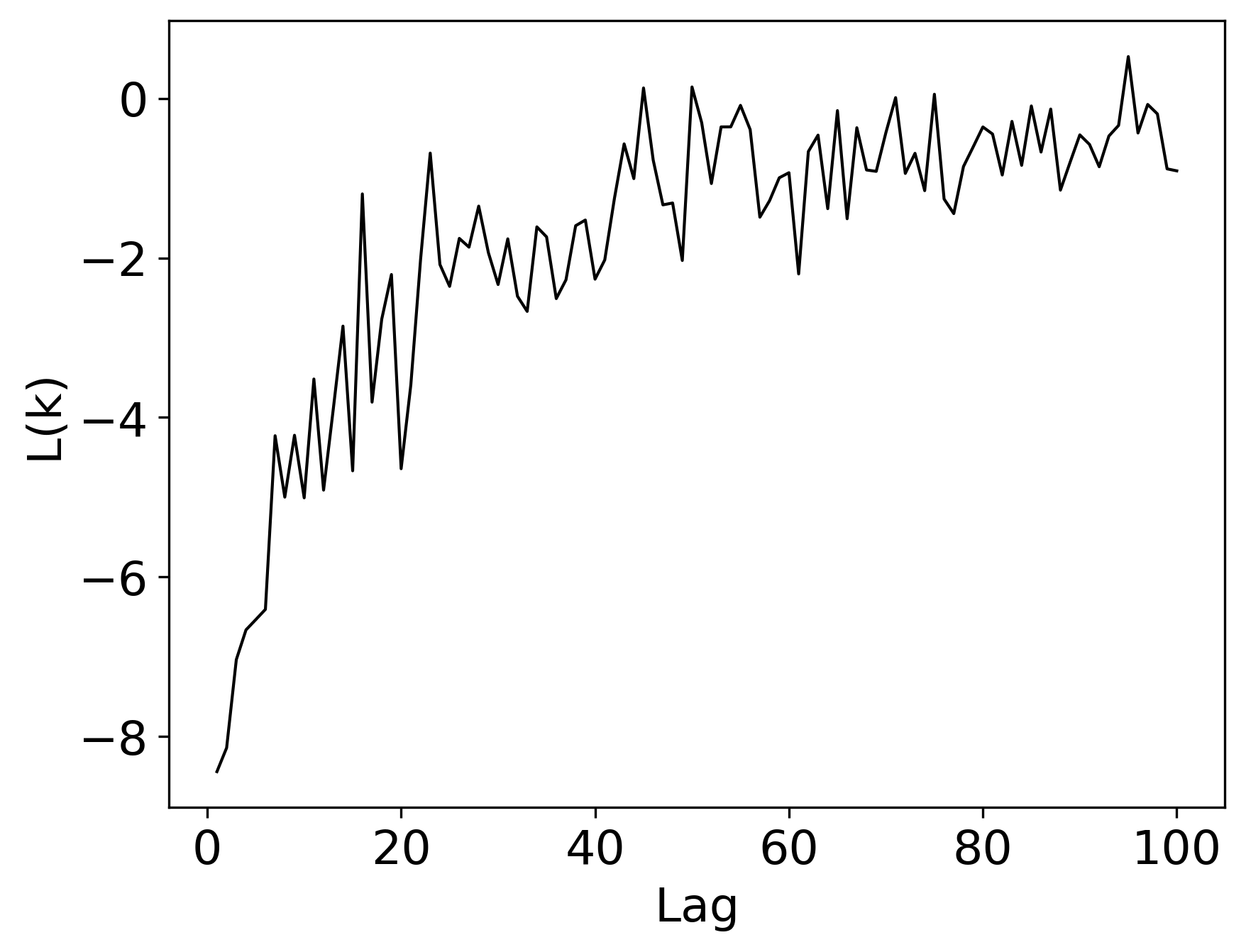}
    \caption{Leverage effect}
   
  \end{subfigure}

  \caption{Stylized facts observed in the stock prices of S\&P 500 firms:
(a) Heavy-tailed distribution, as evidenced by the power-law decay in the tails of the return distribution.
(b) Volatility clustering, reflected by the slow decay of the autocorrelation function of absolute returns.
(c) Leverage effect, shown as a negative lead–lag correlation between returns and future squared returns.}
  \label{fig:ori-plot}
\end{figure}

We evaluate how different forward SDEs (VE, VP, GBM) combined with three commonly used noise variance schedules, affect the preservation of key stylized facts in synthetic financial time series. Specifically, we generate 120 synthetic time series, each of length $L$ = 2048, under every experimental configuration, and compute three statistical measures to assess (i) distributional properties, (ii) volatility clustering, and (iii) the leverage effect. For the computation of those quantities, we follow the method proposed in~\cite{takahashi2019modeling}. Figure~\ref{fig:ori-plot} reports benchmark values derived from real-world financial data for comparison.

During training, we use a minibatch size of 64 and train the score network for 1{,}000 epochs. Following the setting in~\cite{song2020score}, we set the forward time horizon to $T = 1$ and discretize the reverse diffusion process into $N = 2000$ steps. 

For the noise schedule, we focus on three representative noise schedules. The linear schedule increases variance at a constant rate, with  
\[
\sigma_t^2 = \sigma_{\min}^2 + t\,(\sigma_{\max}^2 - \sigma_{\min}^2),
\]  
thereby spreading noise growth evenly over time. The exponential schedule implements a geometric progression of noise strength,  
\[
\sigma_t = \sigma_{\min}\,\Bigl(\tfrac{\sigma_{\max}}{\sigma_{\min}}\Bigr)^t,
\]  
which keeps early timesteps relatively clean and accelerates noise injection as \(t\) approaches 1. Finally, the cosine schedule provides a smooth interpolation between \(\sigma_{\min}\) and \(\sigma_{\max}\) according to  
\[
\sigma_t = \sigma_{\min} + (\sigma_{\max} - \sigma_{\min})\,\frac{1 - \cos(\pi t)}{2},
\]  
mitigating abrupt endpoint transitions and yielding more gradual dynamics. We evaluate each of these schedules in combination with the VE, VP, and GBM SDEs to evaluate their impact on the fidelity of synthetic financial time series. We set the lower and upper bounds to $\sigma_{\min} = 0.01$ and $\sigma_{\max} = 1.0$.

\subsection{Choice the model parameter}\label{subsec:effect}
\begin{figure}[H]
\centering

\begin{minipage}{0.05\textwidth}~\end{minipage}
\begin{minipage}{0.3\textwidth}\centering \textbf{Linear schedule} \end{minipage}
\begin{minipage}{0.3\textwidth}\centering \textbf{Exponential schedule} \end{minipage}
\begin{minipage}{0.3\textwidth}\centering \textbf{Cosine schedule} \end{minipage}

\vspace{0.5em}

\begin{adjustbox}{width=\textwidth}
\begin{minipage}{0.05\textwidth}
  \centering
  \rotatebox{90}{$D_{feat}=16$}
\end{minipage}
\begin{minipage}{0.31\textwidth}
  \includegraphics[width=\linewidth]{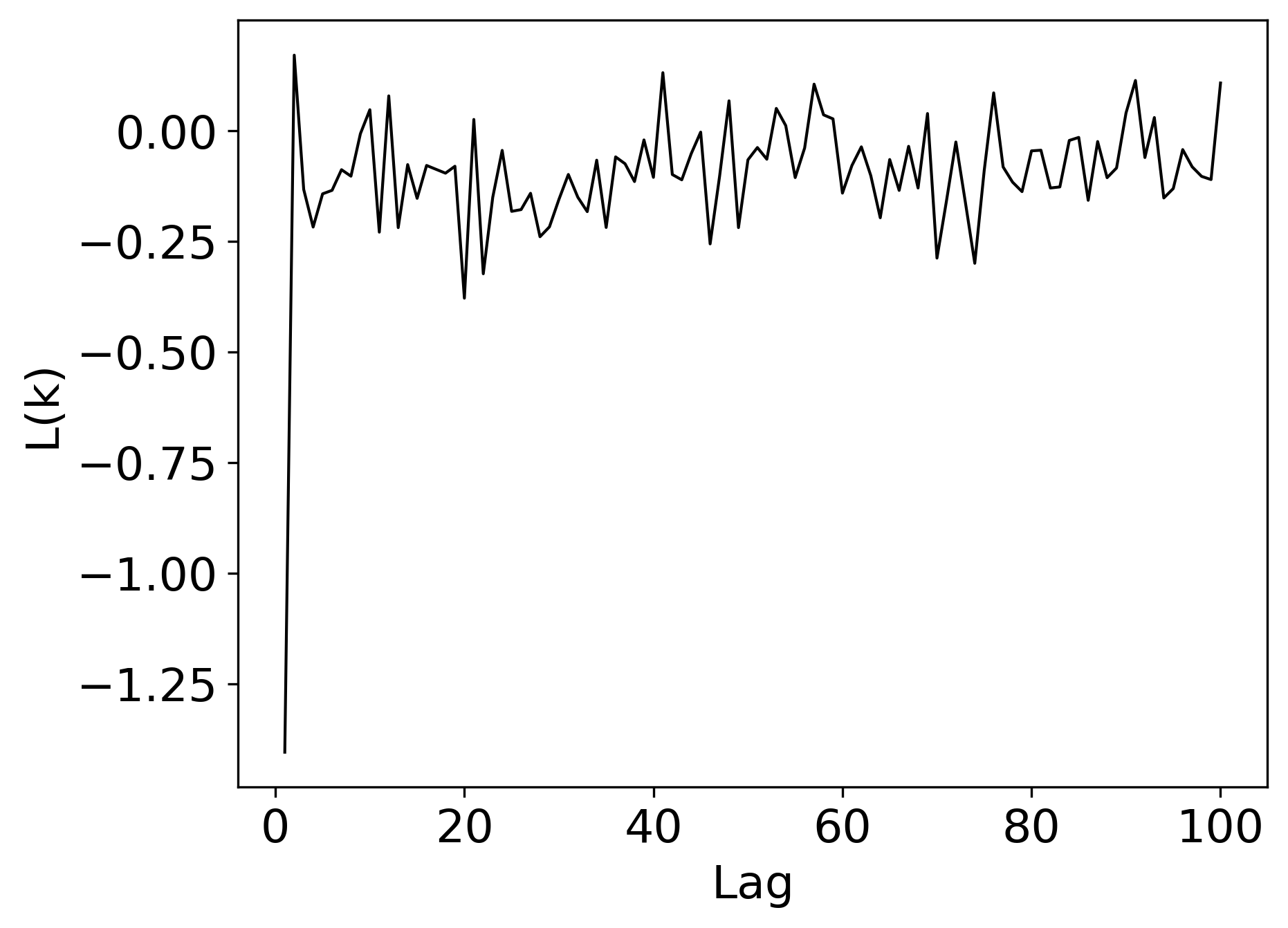}
\end{minipage}
\begin{minipage}{0.31\textwidth}
  \includegraphics[width=\linewidth]{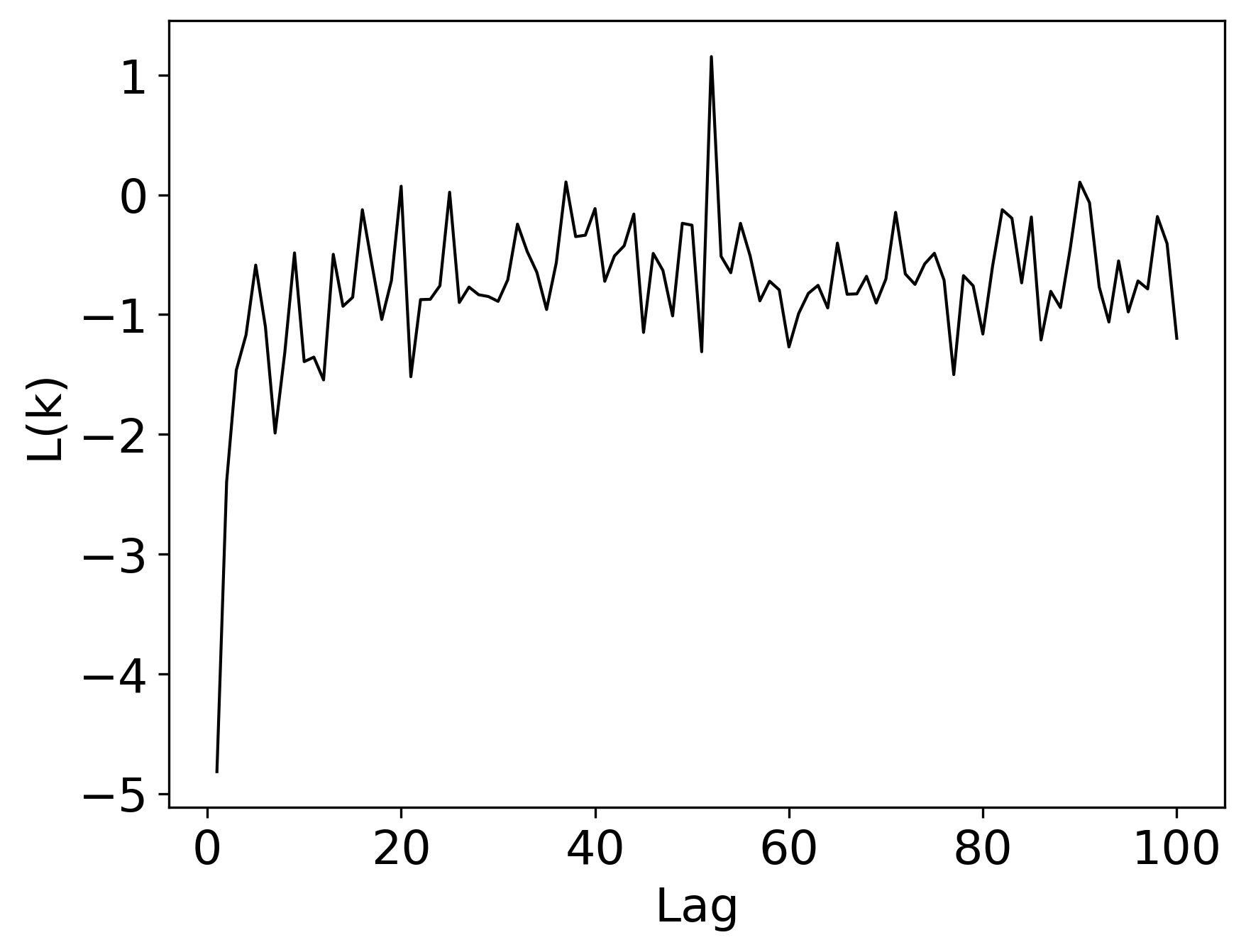}
\end{minipage}
\begin{minipage}{0.31\textwidth}
  \includegraphics[width=\linewidth]{figs/BS_cosine_plot_16/generated_metrics_leverage_effect.png}
\end{minipage}
\end{adjustbox}

\vspace{0.2em}

\begin{adjustbox}{width=\textwidth}
\begin{minipage}{0.05\textwidth}
  \centering
  \rotatebox{90}{\textbf{$D_{feat}=32$}}
\end{minipage}
\begin{minipage}{0.31\textwidth}
  \includegraphics[width=\linewidth]{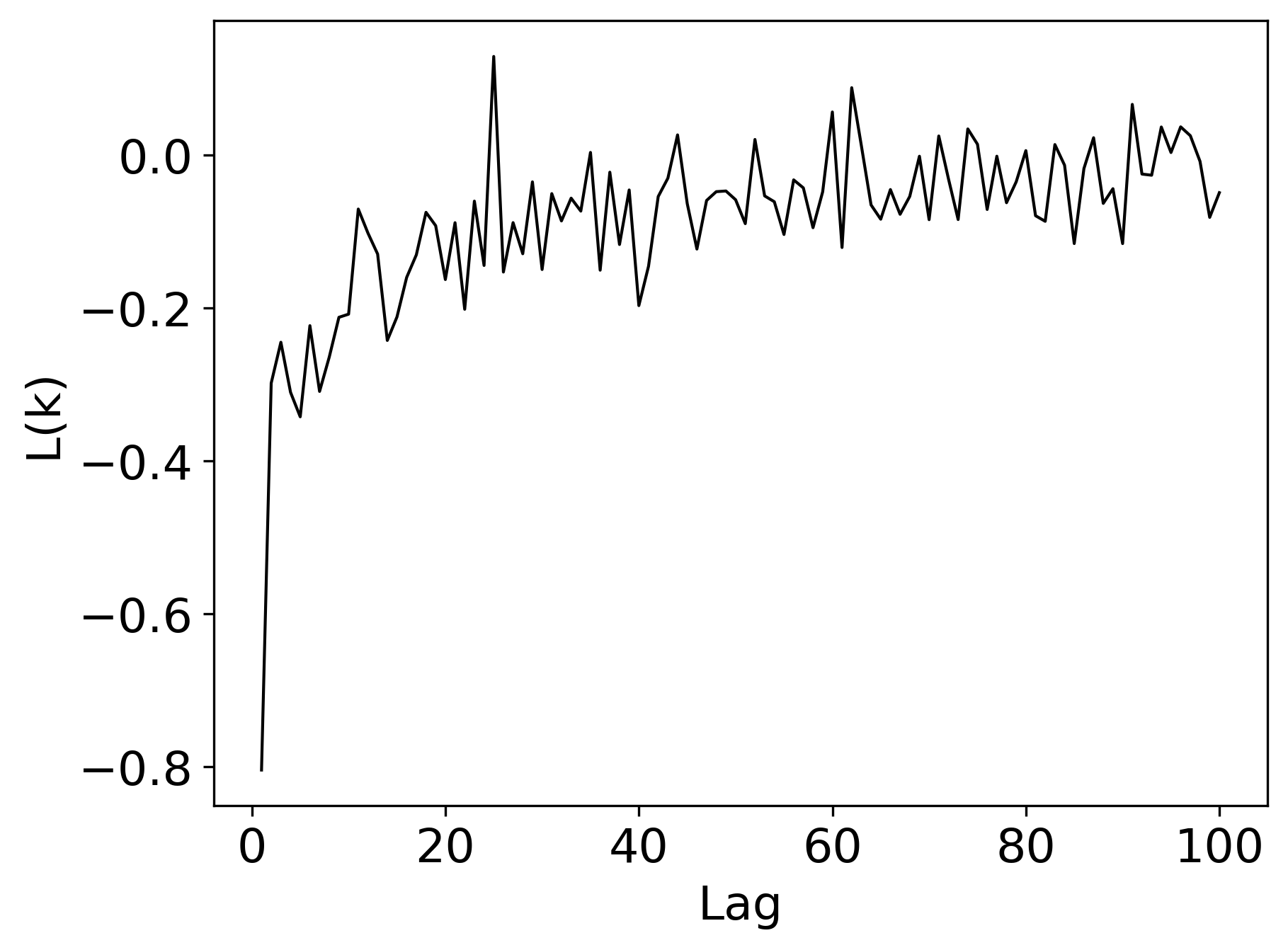}
\end{minipage}
\begin{minipage}{0.31\textwidth}
  \includegraphics[width=\linewidth]{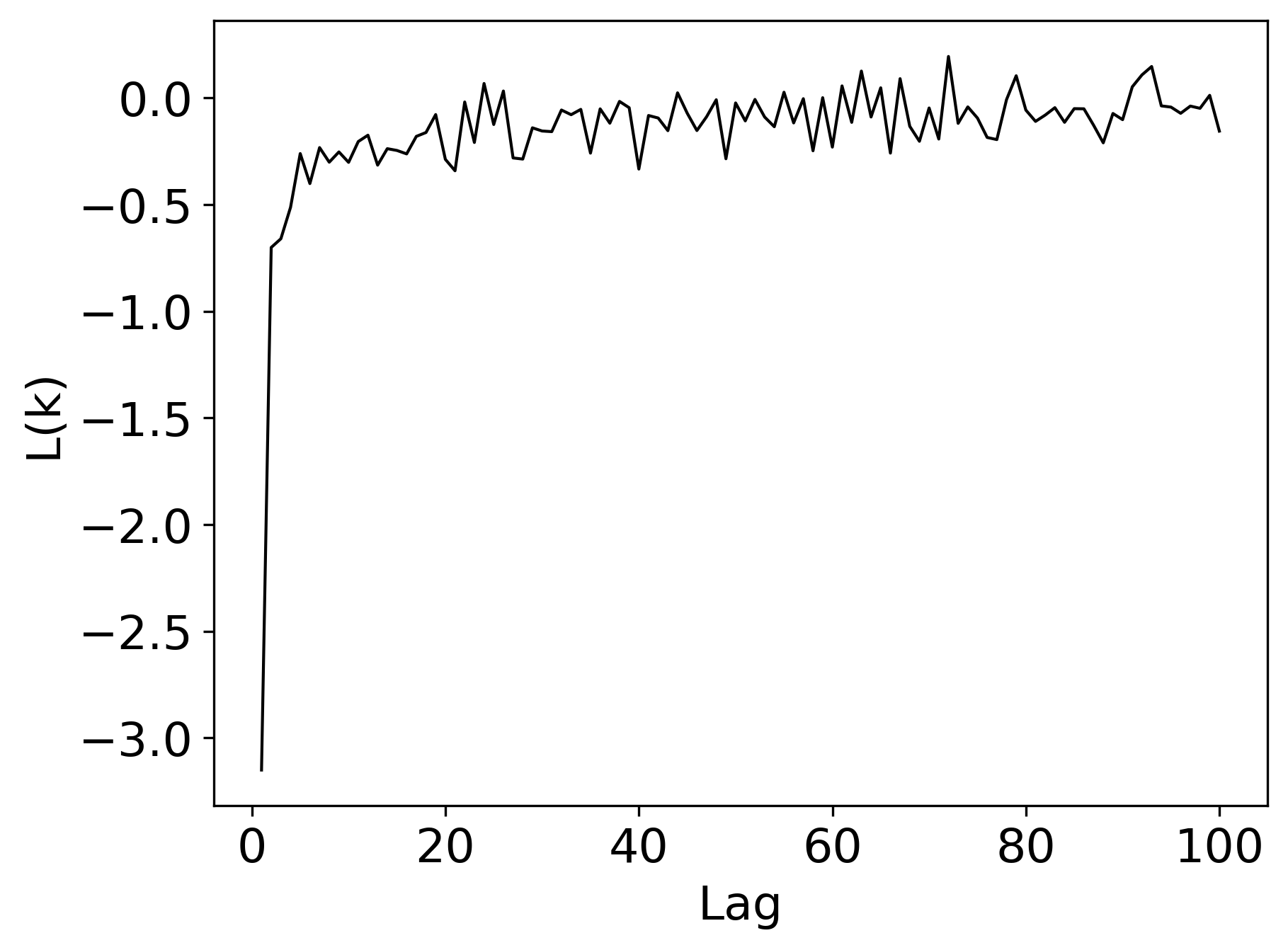}
\end{minipage}
\begin{minipage}{0.31\textwidth}
  \includegraphics[width=\linewidth]{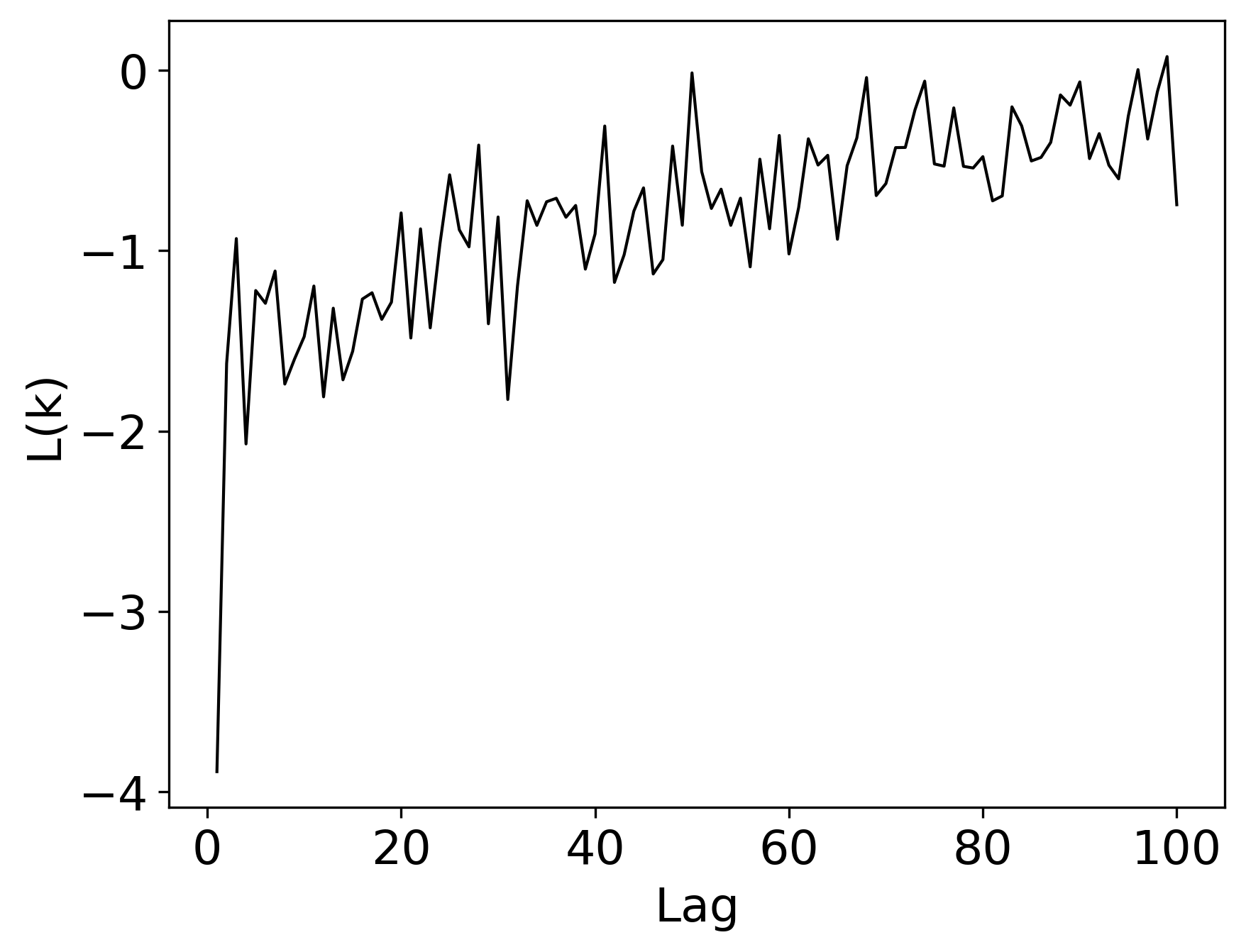}
\end{minipage}
\end{adjustbox}

\begin{adjustbox}{width=\textwidth}
\begin{minipage}{0.05\textwidth}
  \centering
  \rotatebox{90}{$D_{feat}=64$}
\end{minipage}
\begin{minipage}{0.31\textwidth}
  \includegraphics[width=\linewidth]{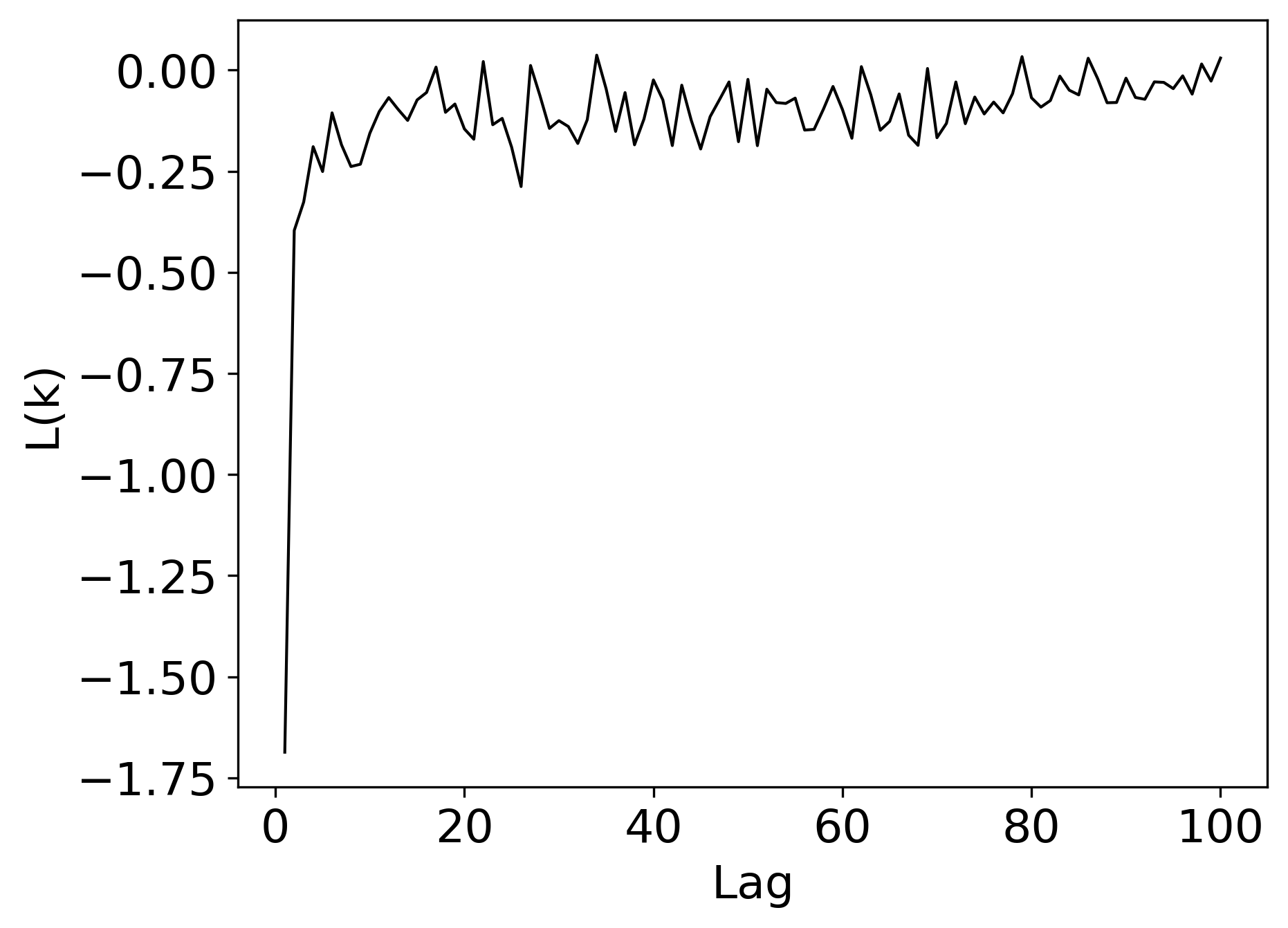}
\end{minipage}
\begin{minipage}{0.31\textwidth}
  \includegraphics[width=\linewidth]{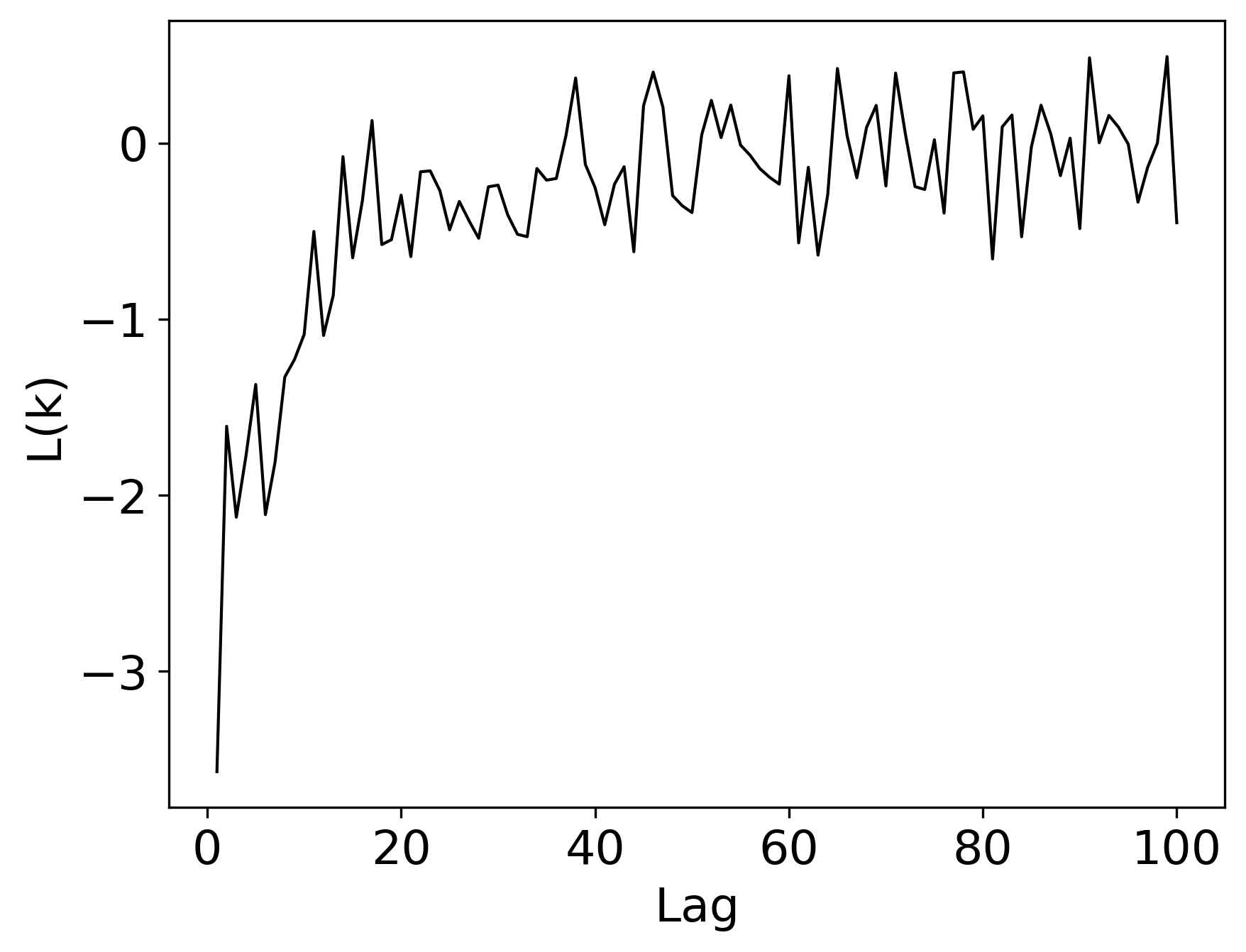}
\end{minipage}
\begin{minipage}{0.31\textwidth}
  \includegraphics[width=\linewidth]{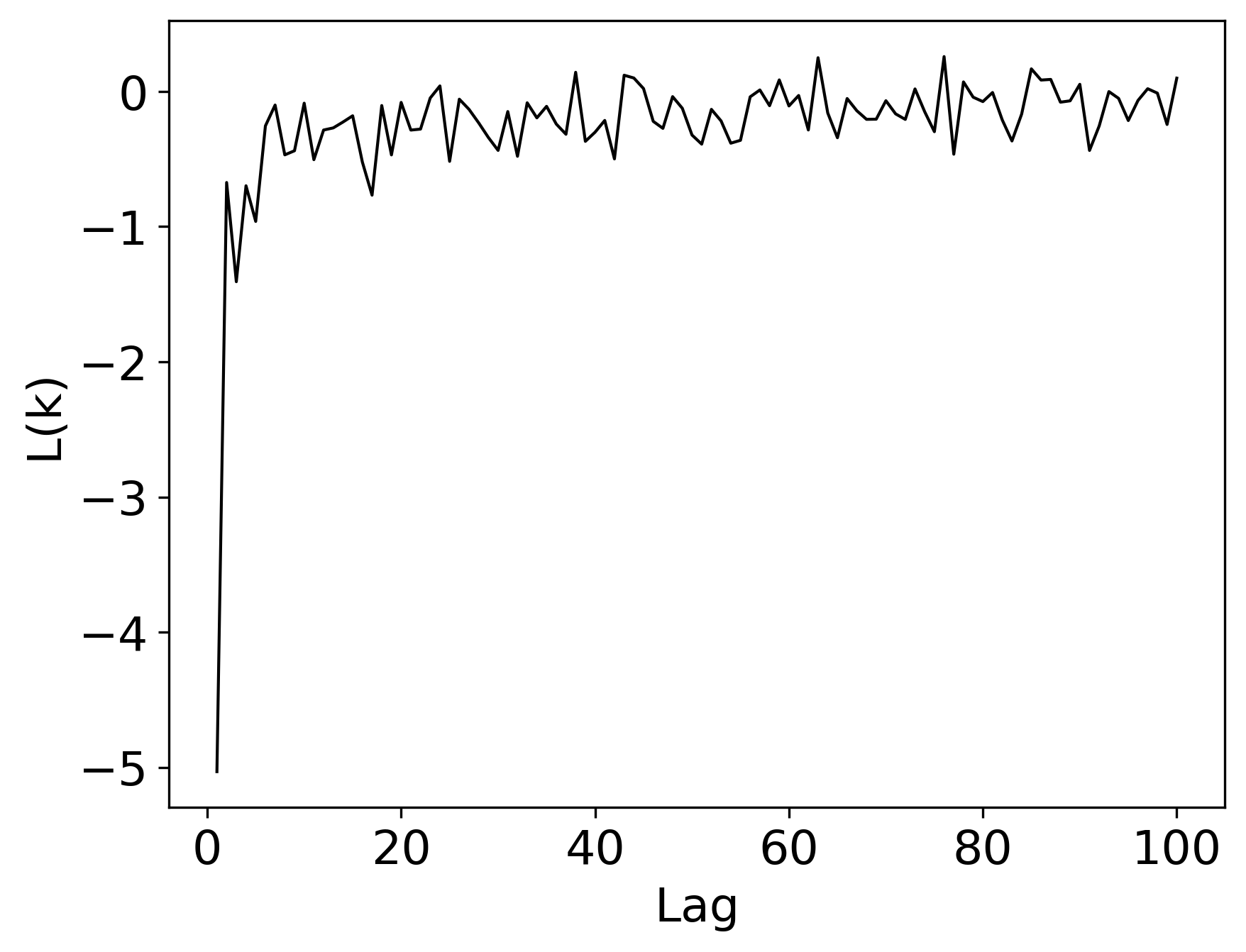}
\end{minipage}
\end{adjustbox}

\caption{
Leverage effect across three different neural network configurations (rows) and noise schedules (columns). Top: 64 convolutional channels, 128-dimensional diffusion-step embedding, and 16-dimensional feature embedding. Middle: 128 convolutional channels, 256-dimensional diffusion-step embedding, and 32-dimensional feature embedding. Bottom: same as middle, but with a 64-dimensional feature embedding.
}
\label{fig:leverage-effect_16_32}
\end{figure}
To investigate how model capacity influences the ability to capture the leverage effect, we evaluate three configurations of the score network that differ in convolutional depth and embedding dimensionality. The baseline architecture mirrors the original CSDI design, employing 64 convolutional channels, a 128-dimensional diffusion-step embedding, and a 16-dimensional feature embedding. We then explore two enhanced variants: the first uniformly doubles the convolutional and embedding dimensions to 128/256/32, while the second further increases the feature embedding size to 64, yielding a 128/256/64 configuration.

As illustrated in Figure~\ref{fig:leverage-effect_16_32}, increasing model capacity consistently improves the network’s ability to replicate the asymmetric correlation structure between returns and future volatility. Notably, enlarging the feature embedding plays a pivotal role in accurately modeling the delayed volatility response that characterizes the leverage effect. Among the three, the 128/256/64 model shows the most realistic and stable leverage profiles across different noise schedules, without introducing overfitting artifacts or spurious long-lag correlations. Accordingly, we adopt this configuration as the default architecture in all subsequent experiments.

\subsection{Heavy‐Tailed distributions}
\begin{figure}[H]
\centering
\begin{minipage}{0.05\textwidth}~\end{minipage}
\begin{minipage}{0.3\textwidth}\centering \textbf{Linear schedule} \end{minipage}
\begin{minipage}{0.3\textwidth}\centering \textbf{Exponential schedule} \end{minipage}
\begin{minipage}{0.3\textwidth}\centering \textbf{Cosine schedule} \end{minipage}

\vspace{0.5em}

\begin{adjustbox}{width=\textwidth}
\begin{minipage}{0.05\textwidth}
  \centering
  \rotatebox{90}{\textbf{VE SDE}}
\end{minipage}
\begin{minipage}{0.31\textwidth}
  \includegraphics[width=\linewidth]{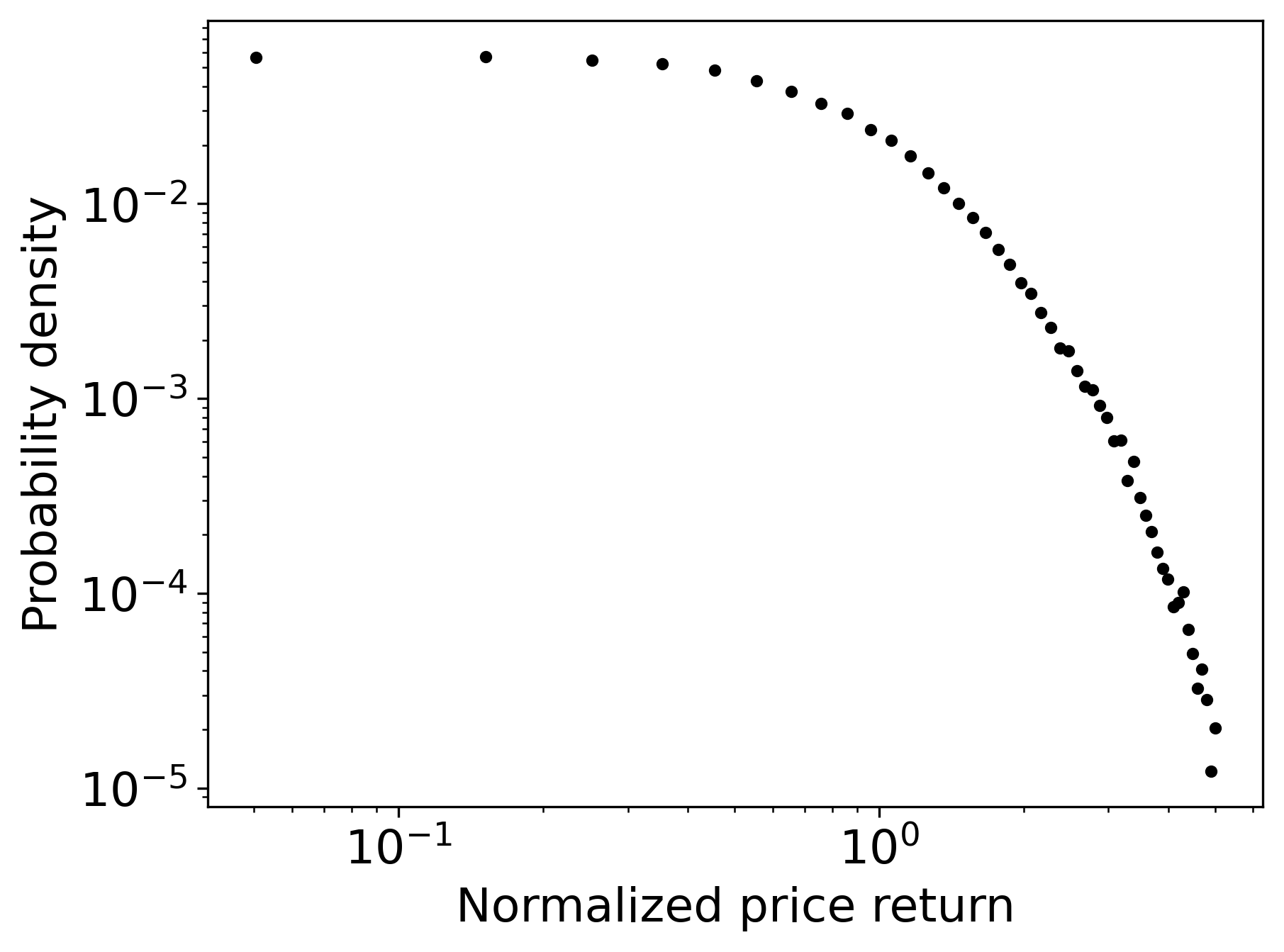}
\end{minipage}
\begin{minipage}{0.31\textwidth}
  \includegraphics[width=\linewidth]{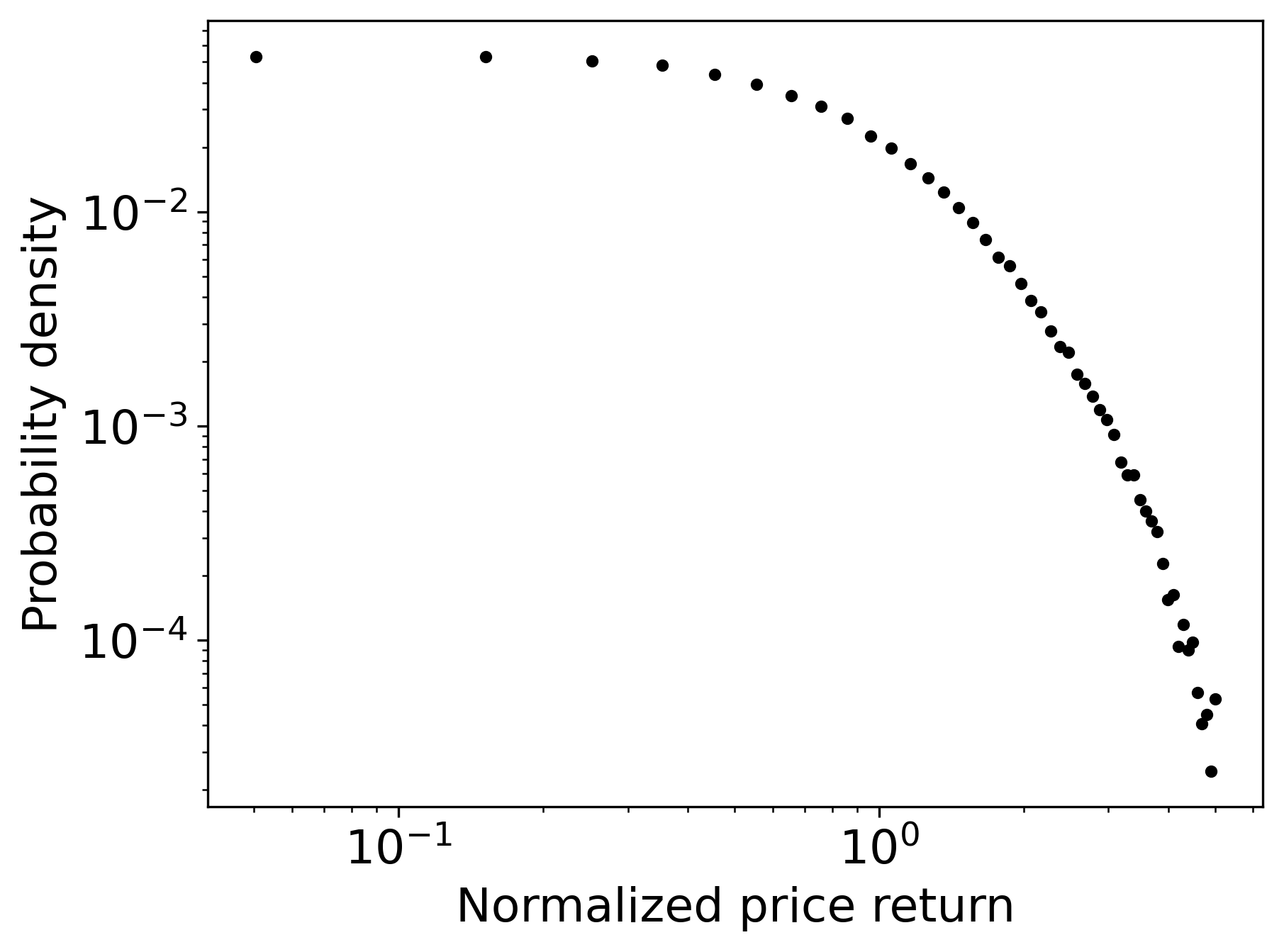}
\end{minipage}
\begin{minipage}{0.31\textwidth}
  \includegraphics[width=\linewidth]{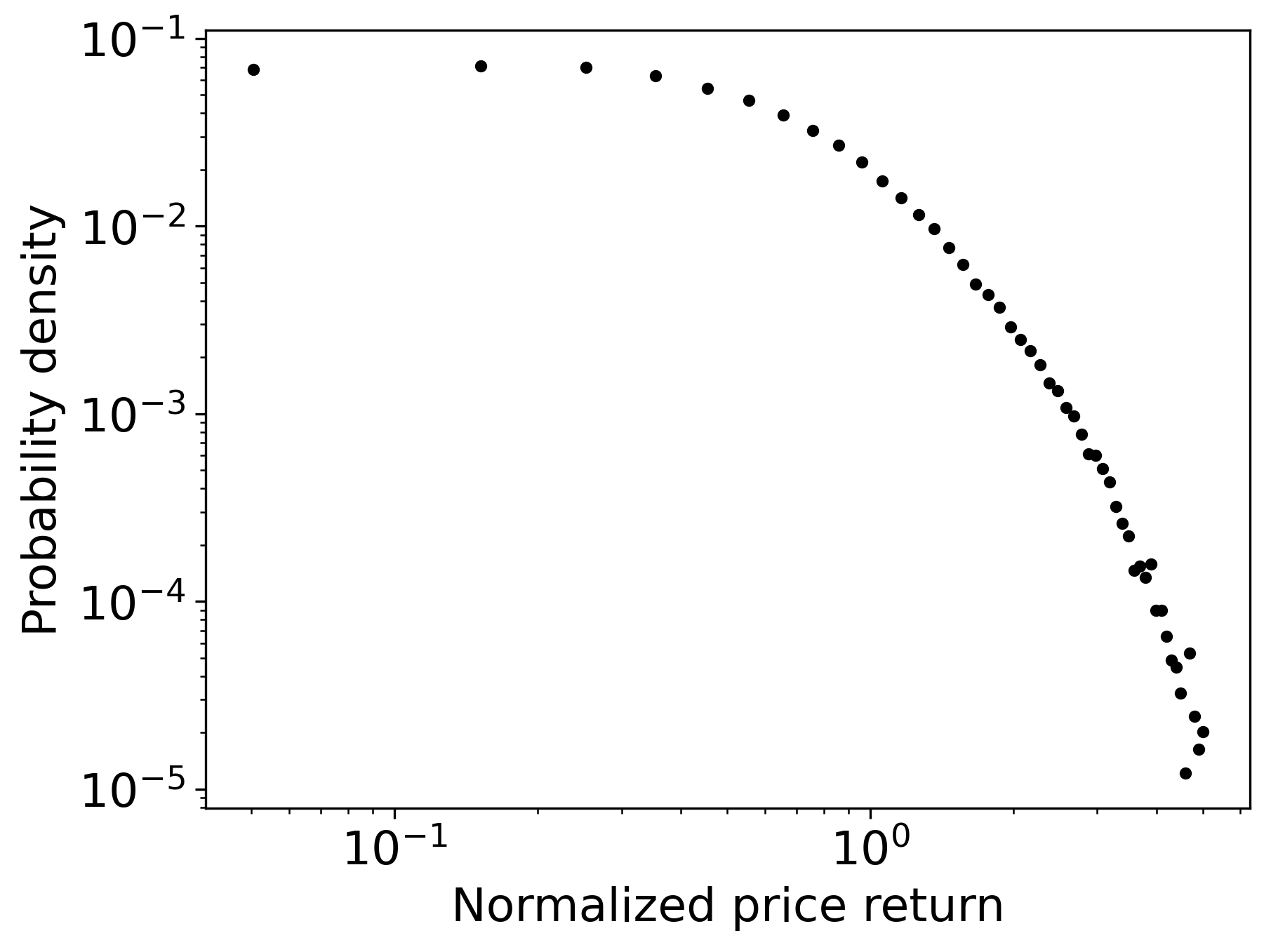}
\end{minipage}
\end{adjustbox}

\vspace{0.2em}

\begin{adjustbox}{width=\textwidth}
\begin{minipage}{0.05\textwidth}
  \centering
  \rotatebox{90}{\textbf{VP SDE}}
\end{minipage}
\begin{minipage}{0.31\textwidth}
  \includegraphics[width=\linewidth]{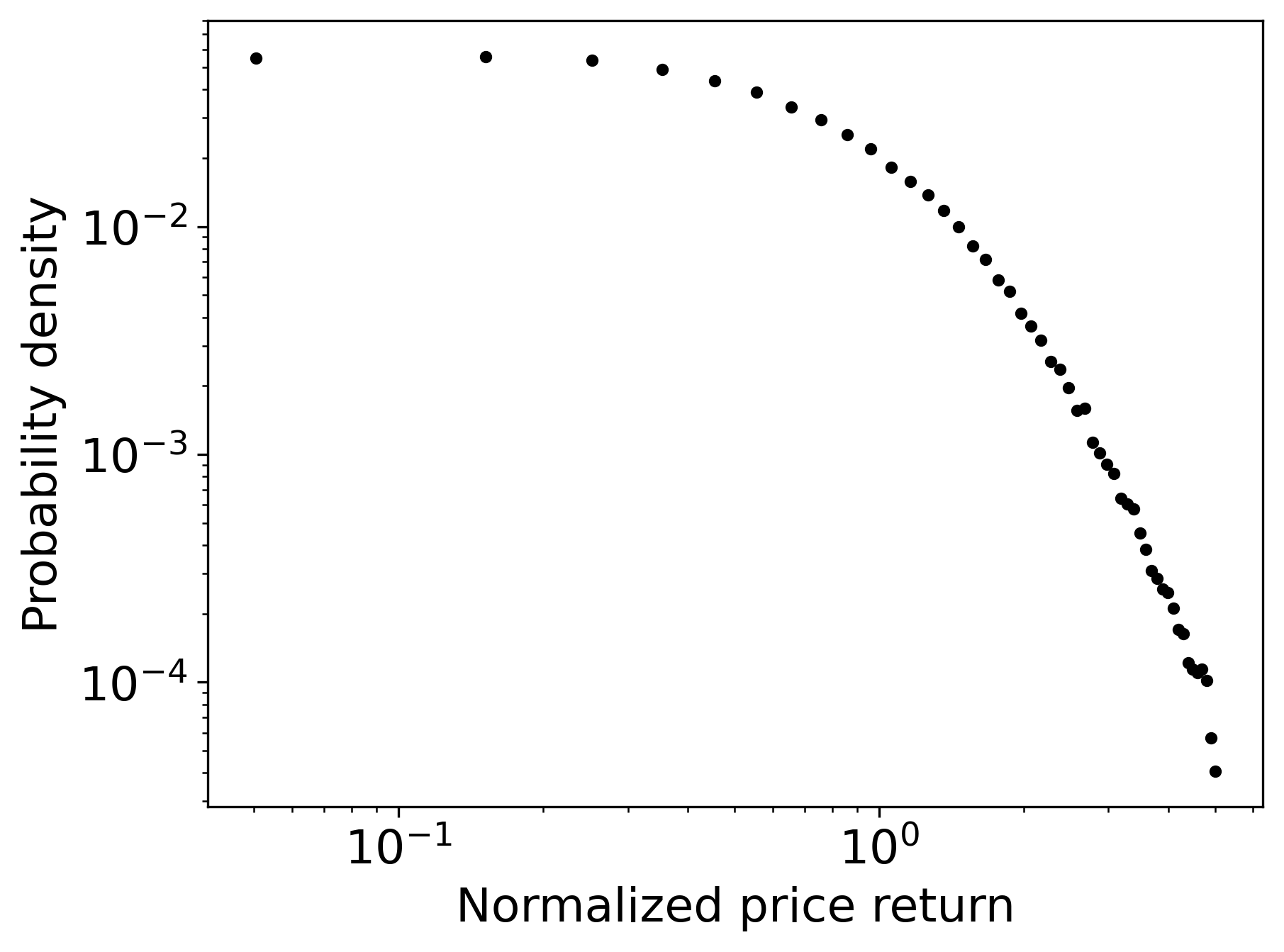}
\end{minipage}
\begin{minipage}{0.31\textwidth}
  \includegraphics[width=\linewidth]{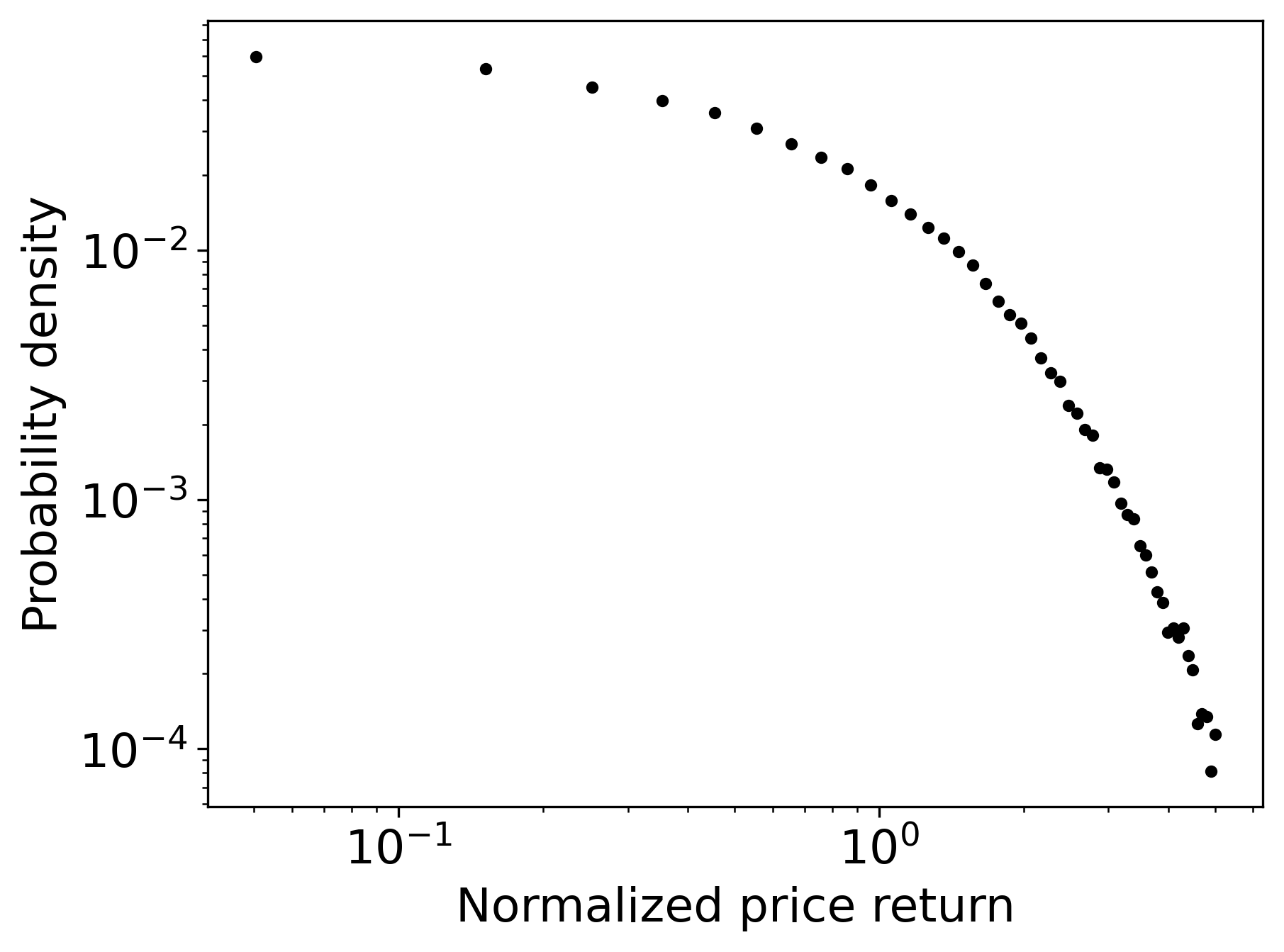}
\end{minipage}
\begin{minipage}{0.31\textwidth}
  \includegraphics[width=\linewidth]{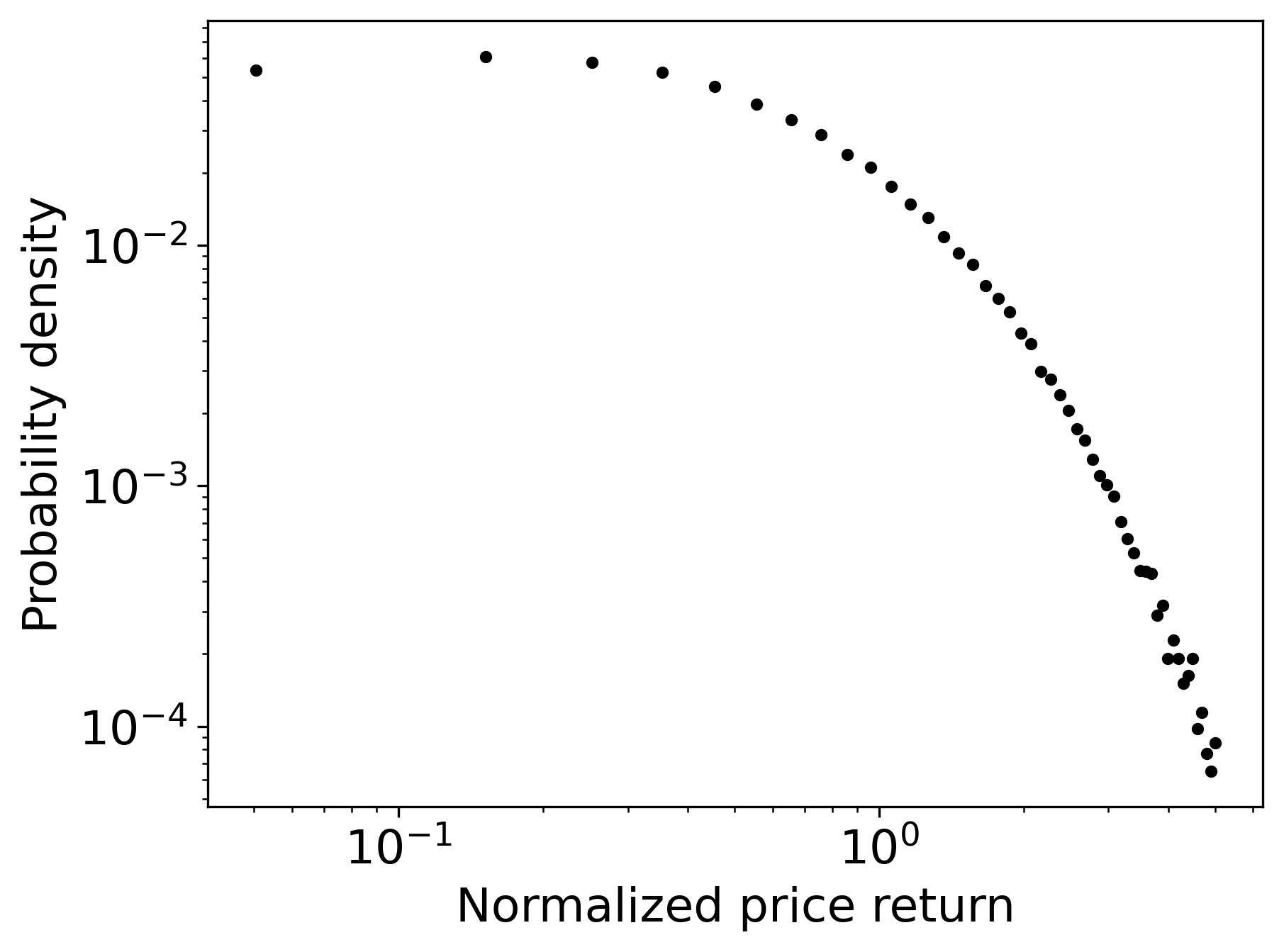}
\end{minipage}
\end{adjustbox}

\vspace{0.2em}

\begin{adjustbox}{width=\textwidth}
\begin{minipage}{0.05\textwidth}
  \centering
  \rotatebox{90}{\textbf{GBM SDE}}
\end{minipage}
\begin{minipage}{0.31\textwidth}
  \includegraphics[width=\linewidth]{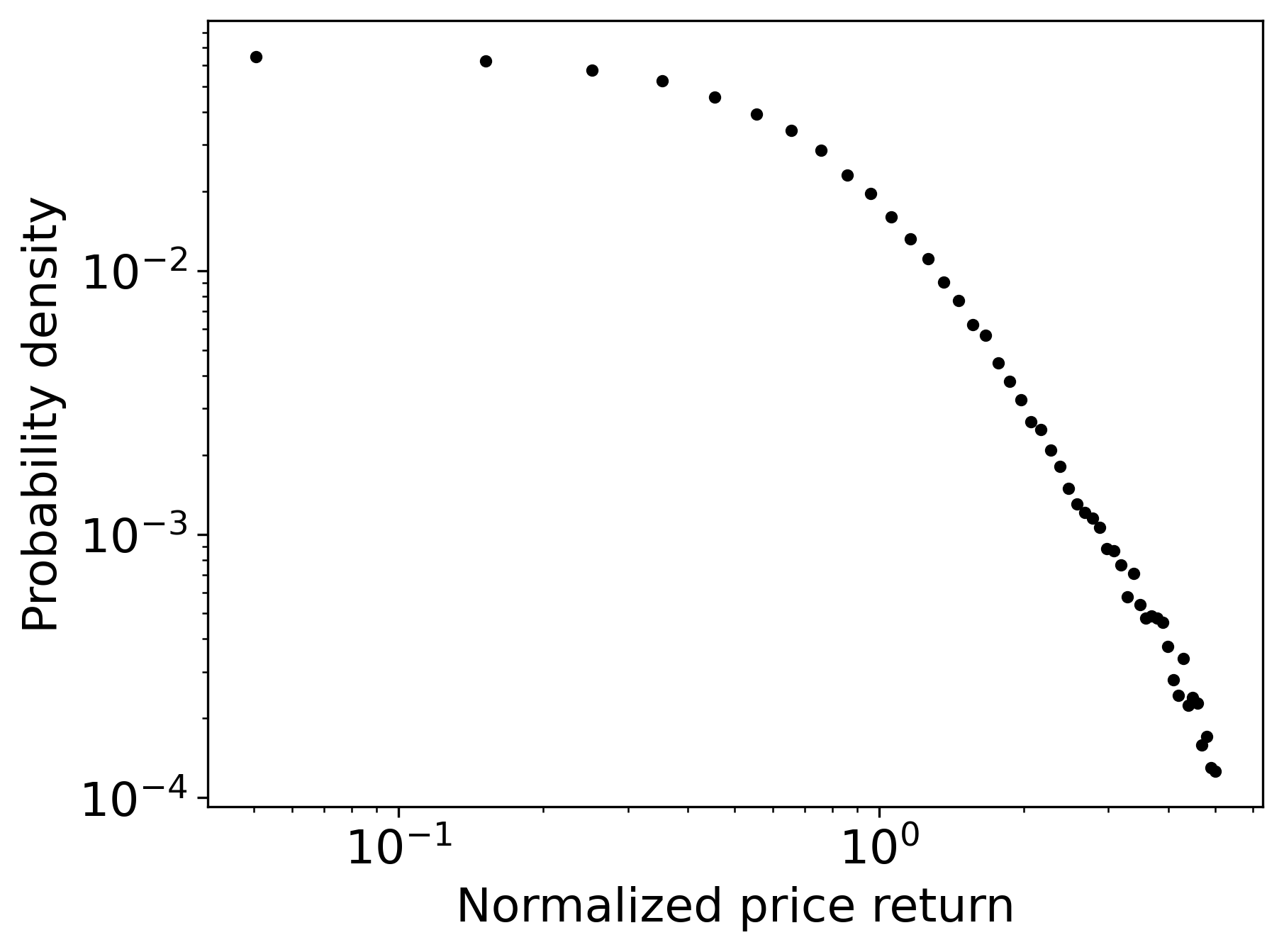}
\end{minipage}
\begin{minipage}{0.31\textwidth}
  \includegraphics[width=\linewidth]{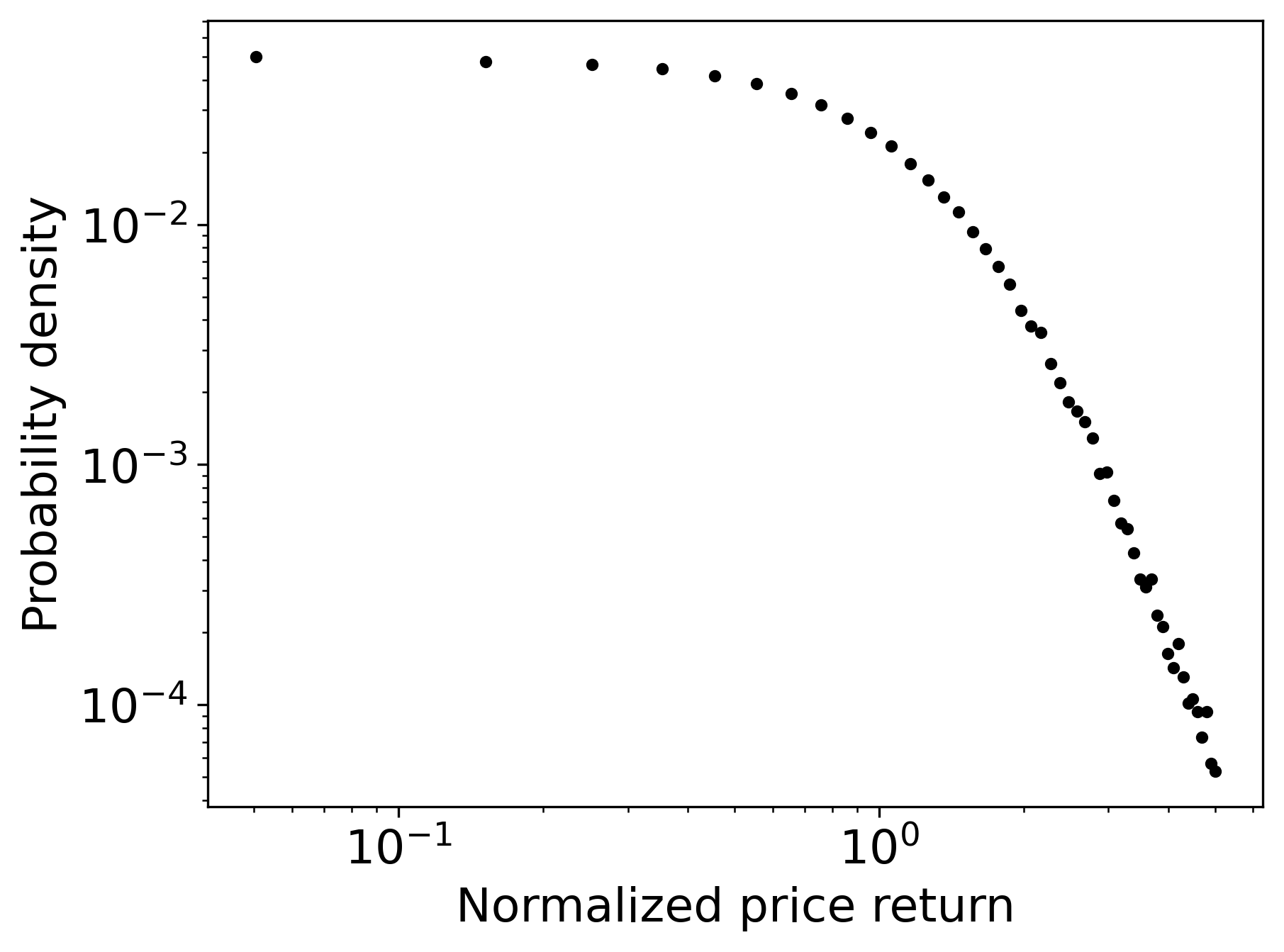}
\end{minipage}
\begin{minipage}{0.31\textwidth}
  \includegraphics[width=\linewidth]{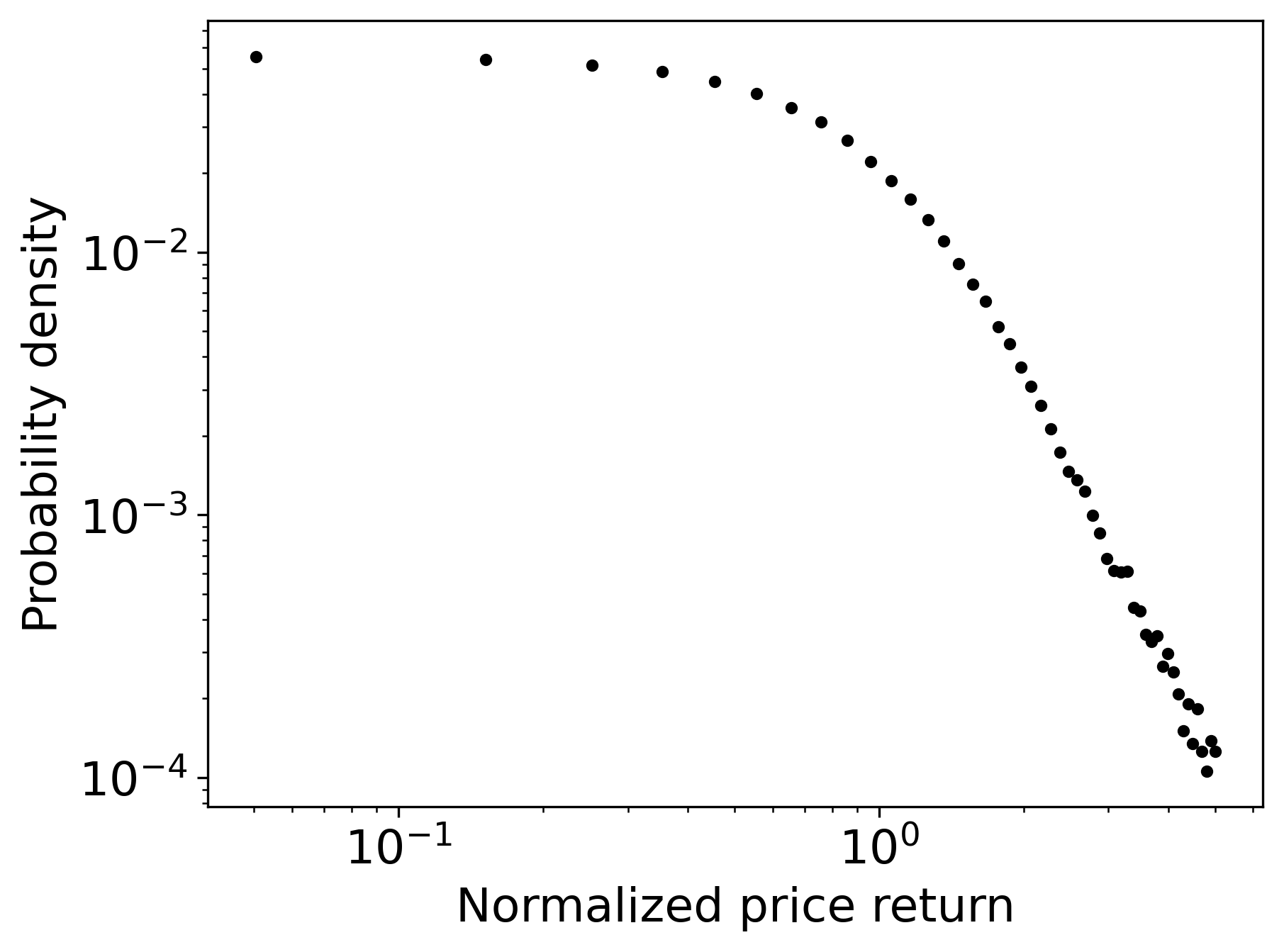}
\end{minipage}
\end{adjustbox}

\caption{
Heavy-tail exponent for three SDE variants (rows) under three noise schedules (columns).  }
\label{fig:heavy-tail}
\end{figure}

We examine the tail behavior of the generated return distributions produced under different forward SDEs and noise schedules by estimating the tail exponent \(\alpha\) of each setup. Figure~\ref{fig:heavy-tail} demonstrates log–log plots of return densities generated by three models: VE, VP, and GBM, under linear, exponential, and cosine noise schedules. For reference, the empirical tail exponent computed from S\&P 500 log-returns is \(\alpha = 4.35\) (Figure~\ref{fig:ori-plot}(a)).

Within the VE SDE framework, the estimated tail exponents are considerably larger than those observed in empirical data: \(\alpha = 8.96\), \(8.49\), and \(4.14\) for linear, exponential, and cosine schedules, respectively. These values indicate light-tailed behavior relative to real market data. This phenomenon can be attributed to the additive nature of the VE SDE in the log-space, which lacks any coupling between volatility and price level. As a result, the diffusion process tends to overly homogenize the magnitude of returns, suppressing the probability of extreme events and thereby overestimating tail decay. The VP SDE, while incorporating a drift term to preserve variance, shares the same additive noise structure and thus suffers from similar limitations. Without a multiplicative scaling mechanism, both VE and VP SDEs fail to induce heteroskedasticity, resulting in lighter tails than observed in real financial markets.

In contrast, the GBM SDE framework yields substantially heavier tails, with \(\alpha = 3.06\), \(4.62\), and \(3.78\) across the three schedules. Notably, the exponential and cosine schedules produce exponents closest to the empirical value. This result highlights the importance of modeling multiplicative noise: in GBM, the diffusion term scales with the price itself, allowing high-return regimes to experience proportionally larger perturbations. This intrinsic coupling introduces heteroskedasticity into the forward process, resulting in a more realistic return distribution with a naturally heavy-tailed structure.

\subsection{Volatility clustering}
\begin{figure}[H]
\centering

\begin{minipage}{0.05\textwidth}~\end{minipage}
\begin{minipage}{0.3\textwidth}\centering \textbf{Linear schedule} \end{minipage}
\begin{minipage}{0.3\textwidth}\centering \textbf{Exponential schedule} \end{minipage}
\begin{minipage}{0.3\textwidth}\centering \textbf{Cosine schedule} \end{minipage}

\vspace{0.5em}

\begin{adjustbox}{width=\textwidth}
\begin{minipage}{0.05\textwidth}
  \centering
  \rotatebox{90}{\textbf{VE SDE}}
\end{minipage}
\begin{minipage}{0.31\textwidth}
  \includegraphics[width=\linewidth]{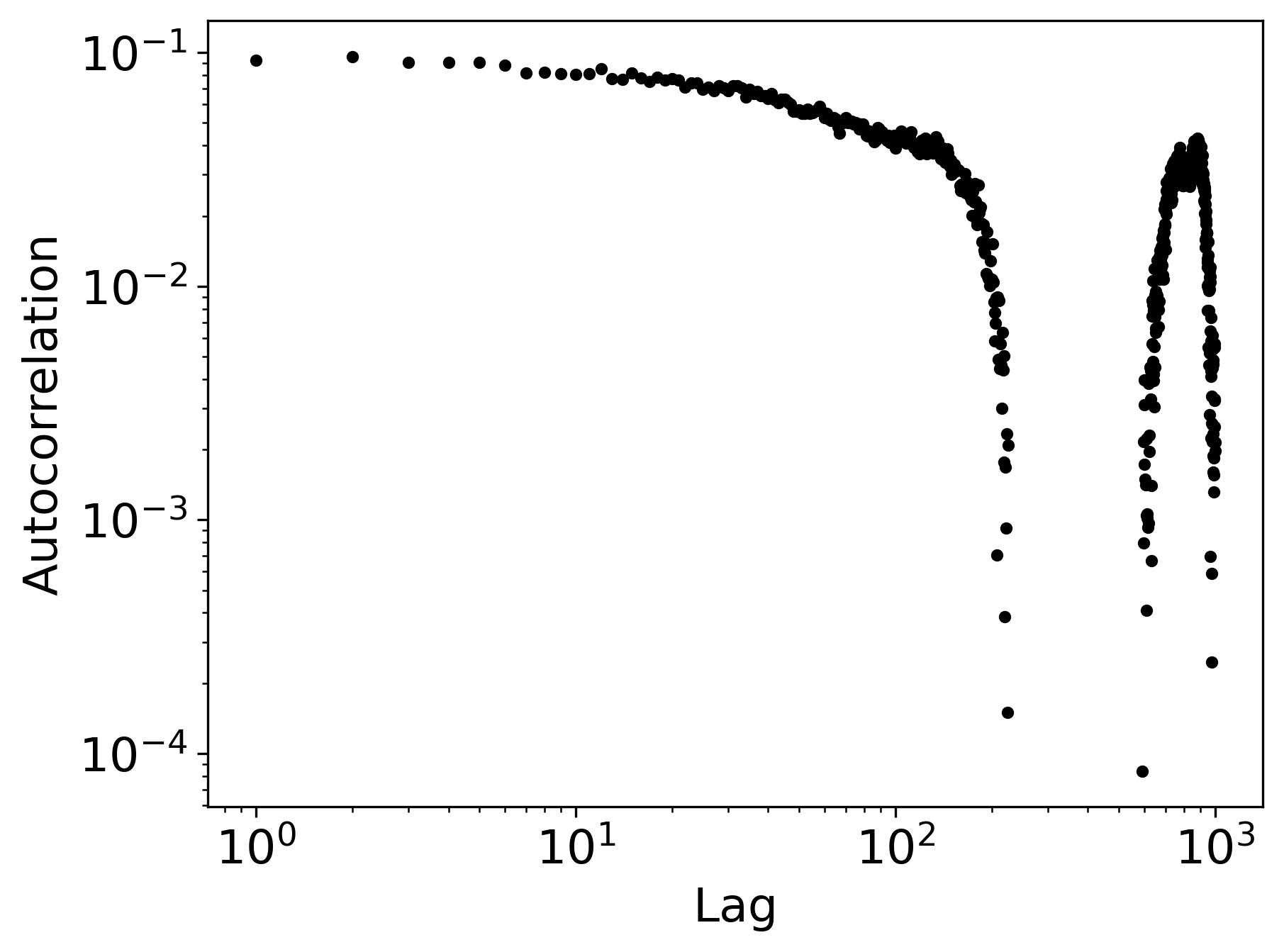}
\end{minipage}
\begin{minipage}{0.31\textwidth}
  \includegraphics[width=\linewidth]{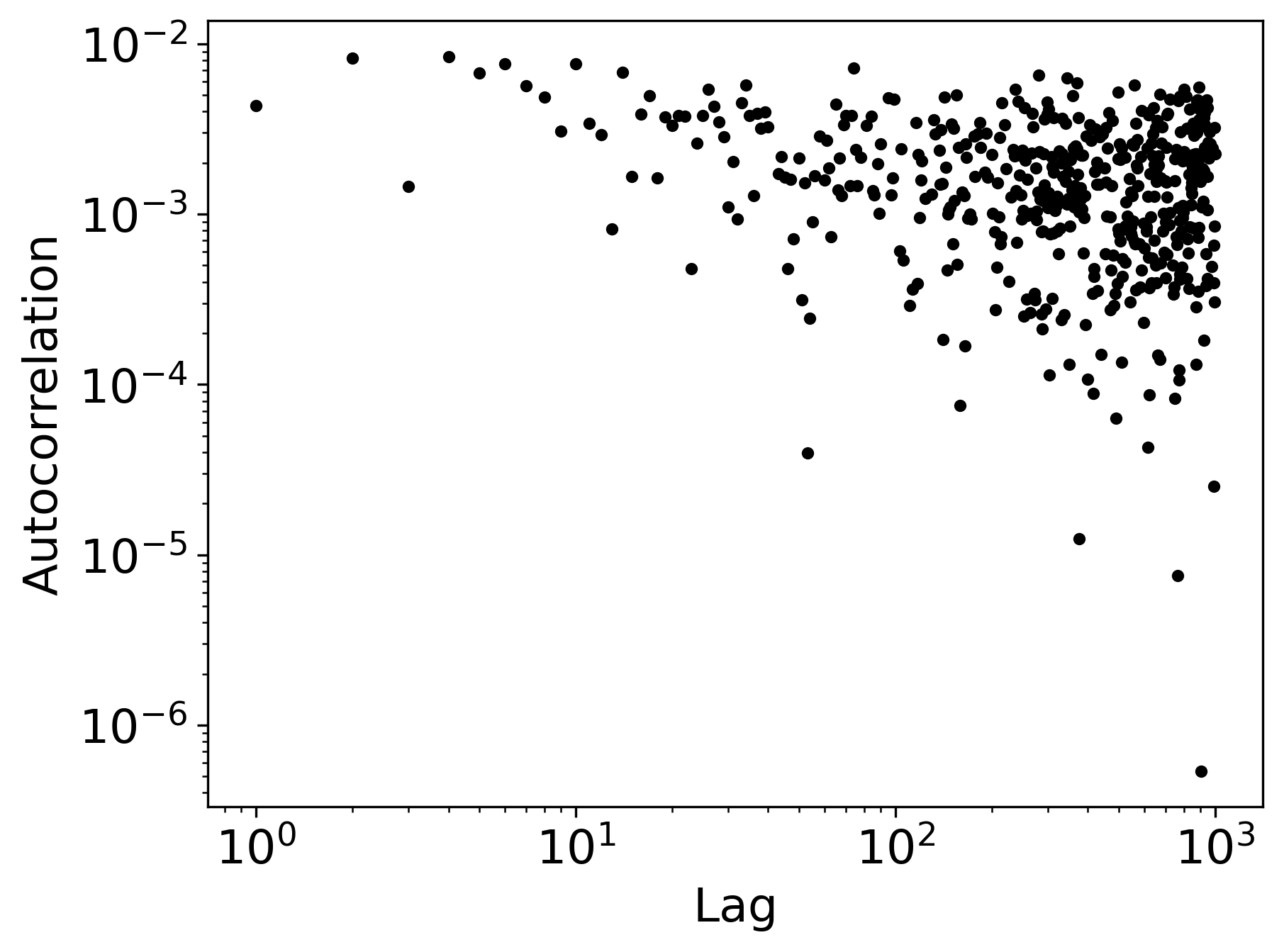}
\end{minipage}
\begin{minipage}{0.31\textwidth}
  \includegraphics[width=\linewidth]{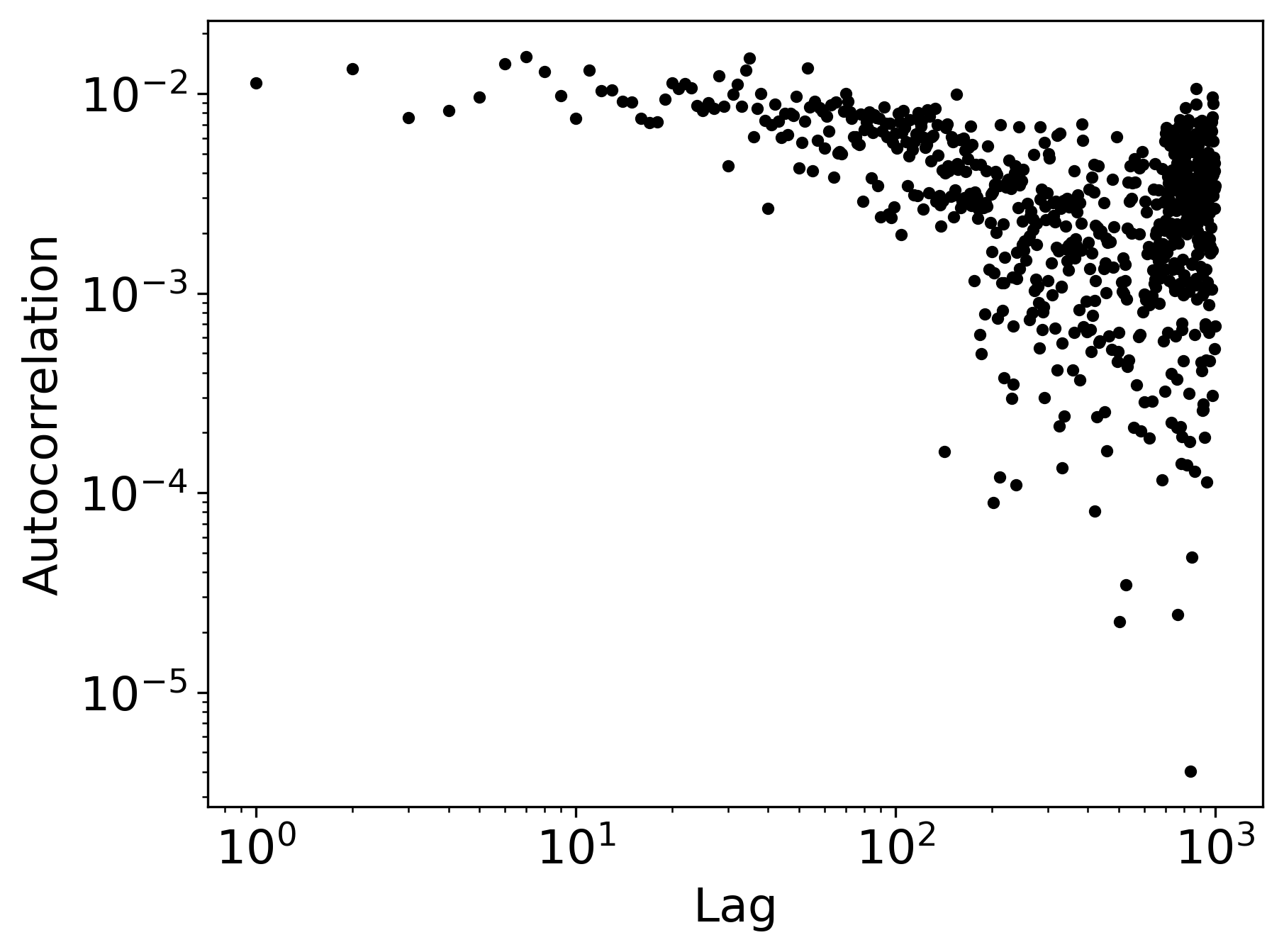}
\end{minipage}
\end{adjustbox}

\vspace{0.2em}

\begin{adjustbox}{width=\textwidth}
\begin{minipage}{0.05\textwidth}
  \centering
  \rotatebox{90}{\textbf{VP SDE}}
\end{minipage}
\begin{minipage}{0.31\textwidth}
  \includegraphics[width=\linewidth]{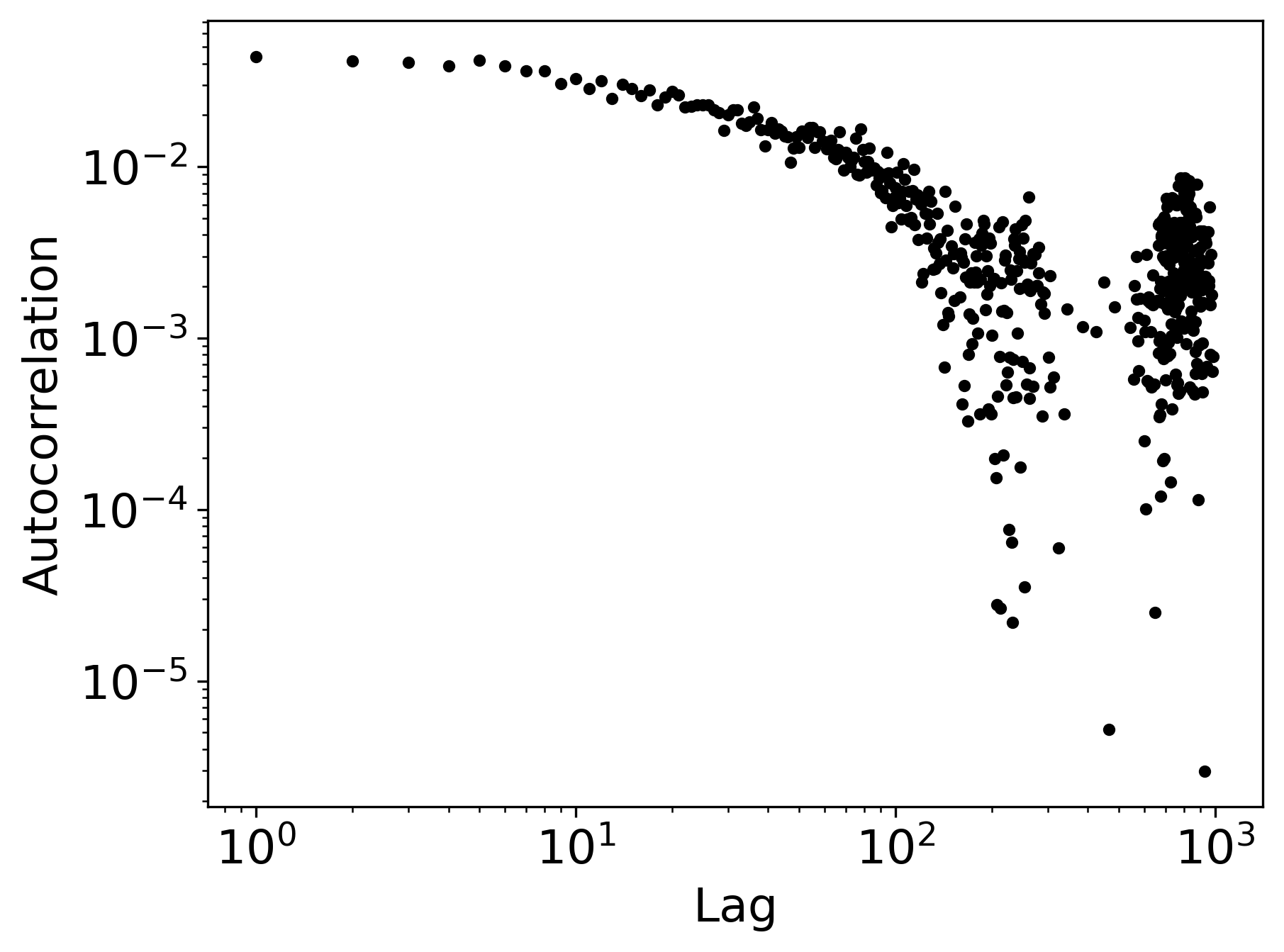}
\end{minipage}
\begin{minipage}{0.31\textwidth}
  \includegraphics[width=\linewidth]{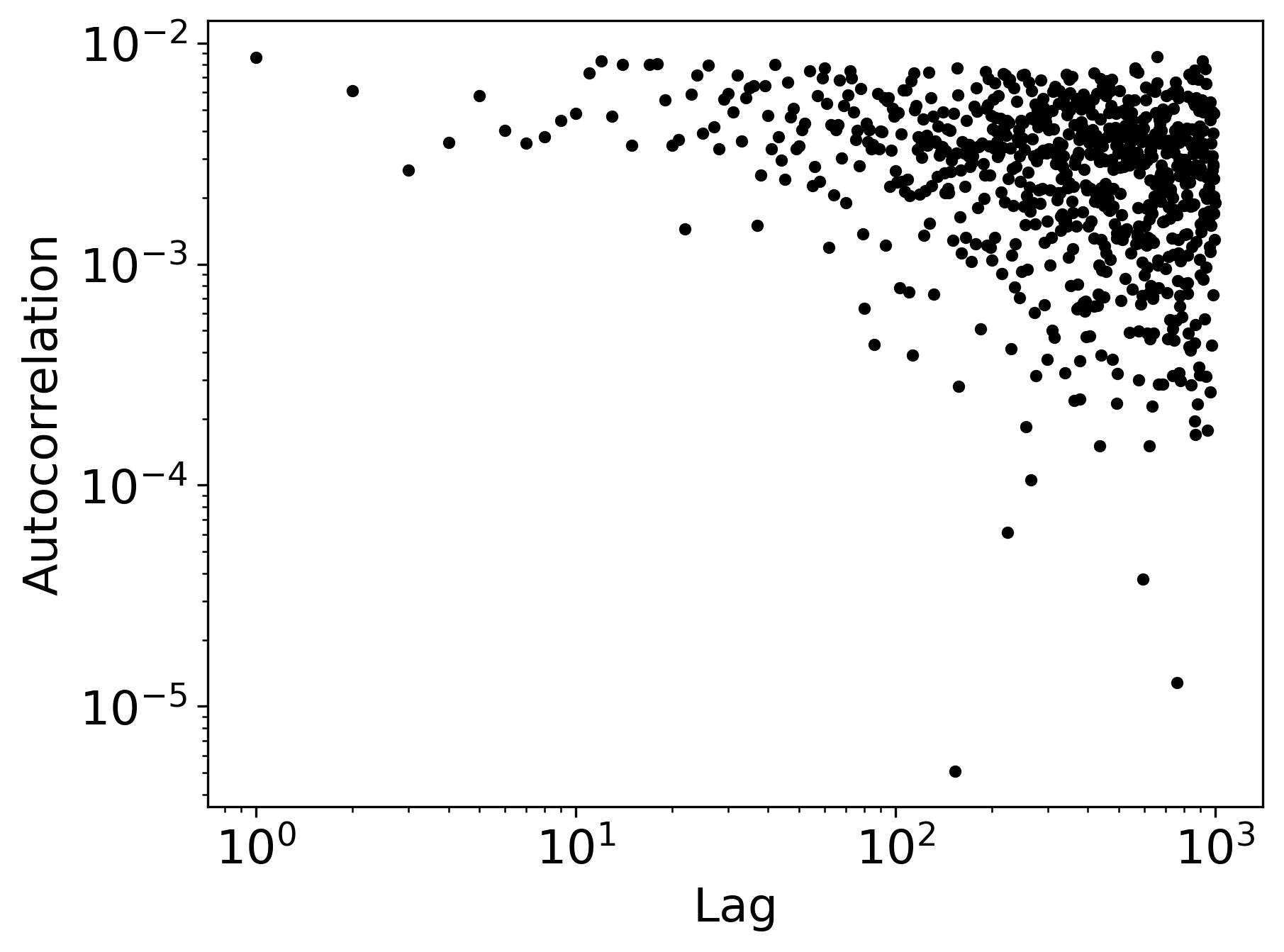}
\end{minipage}
\begin{minipage}{0.31\textwidth}
  \includegraphics[width=\linewidth]{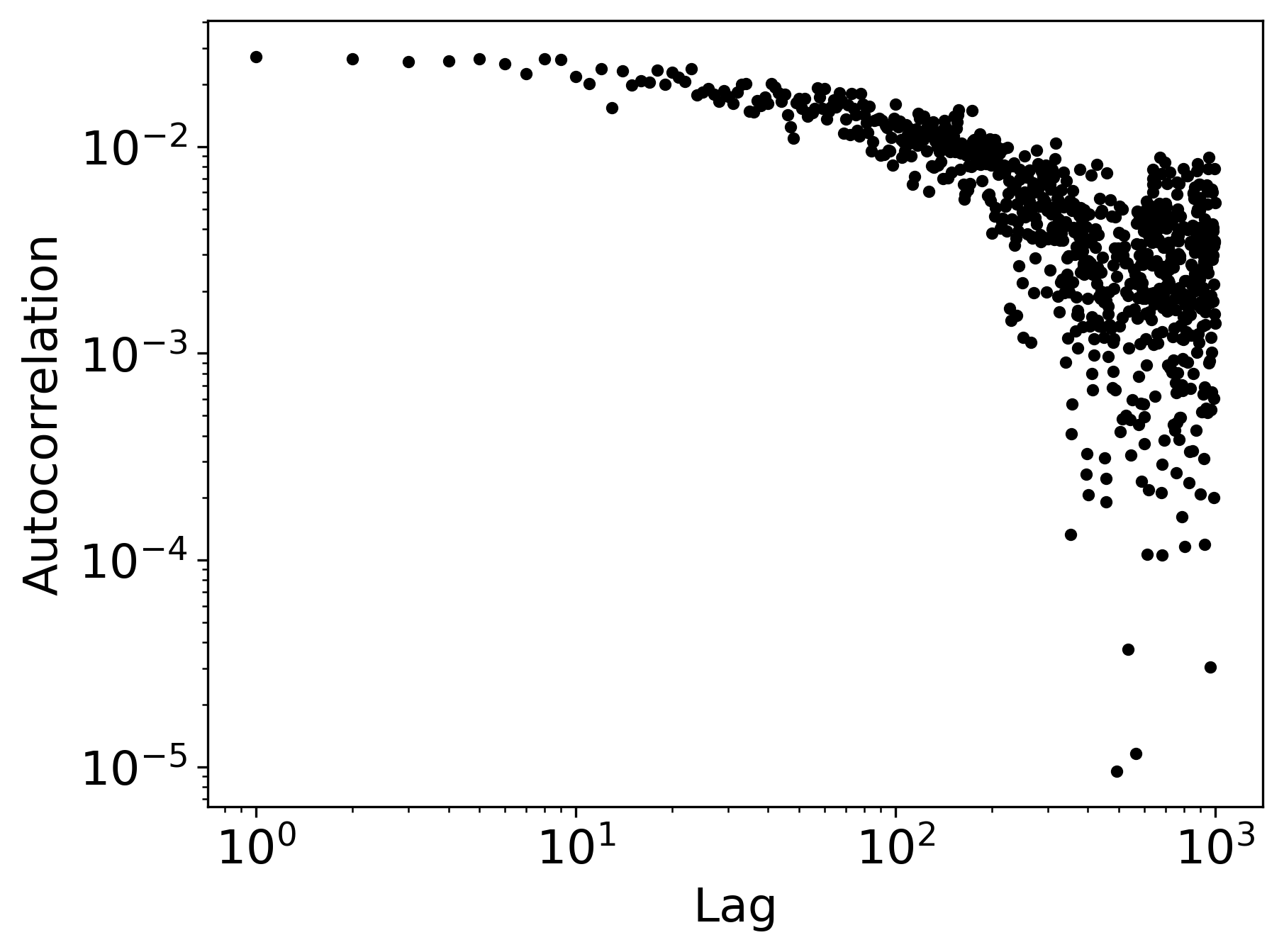}
\end{minipage}
\end{adjustbox}

\vspace{0.2em}

\begin{adjustbox}{width=\textwidth}
\begin{minipage}{0.05\textwidth}
  \centering
  \rotatebox{90}{\textbf{GBM SDE}}
\end{minipage}
\begin{minipage}{0.31\textwidth}
  \includegraphics[width=\linewidth]{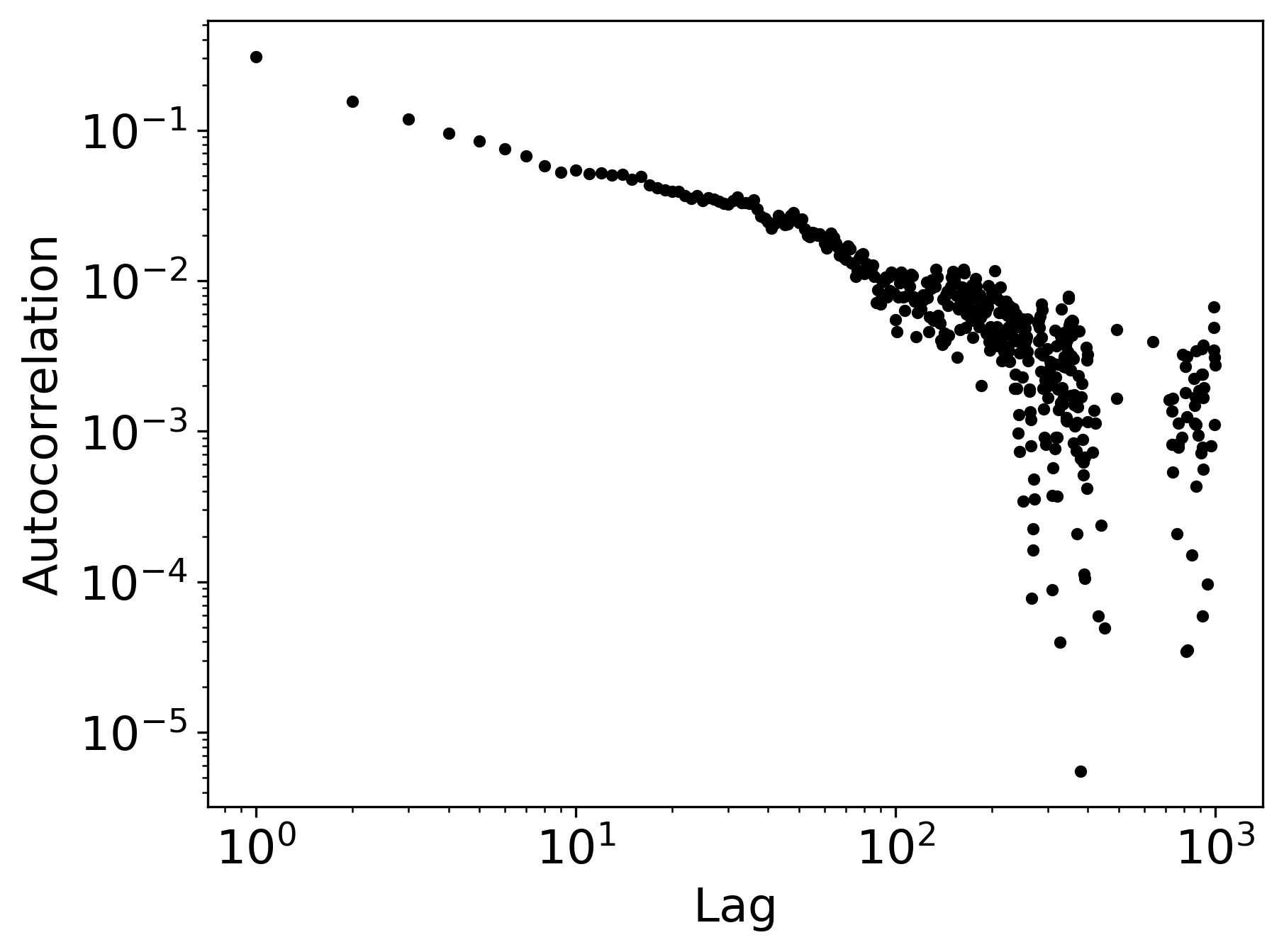}
\end{minipage}
\begin{minipage}{0.31\textwidth}
  \includegraphics[width=\linewidth]{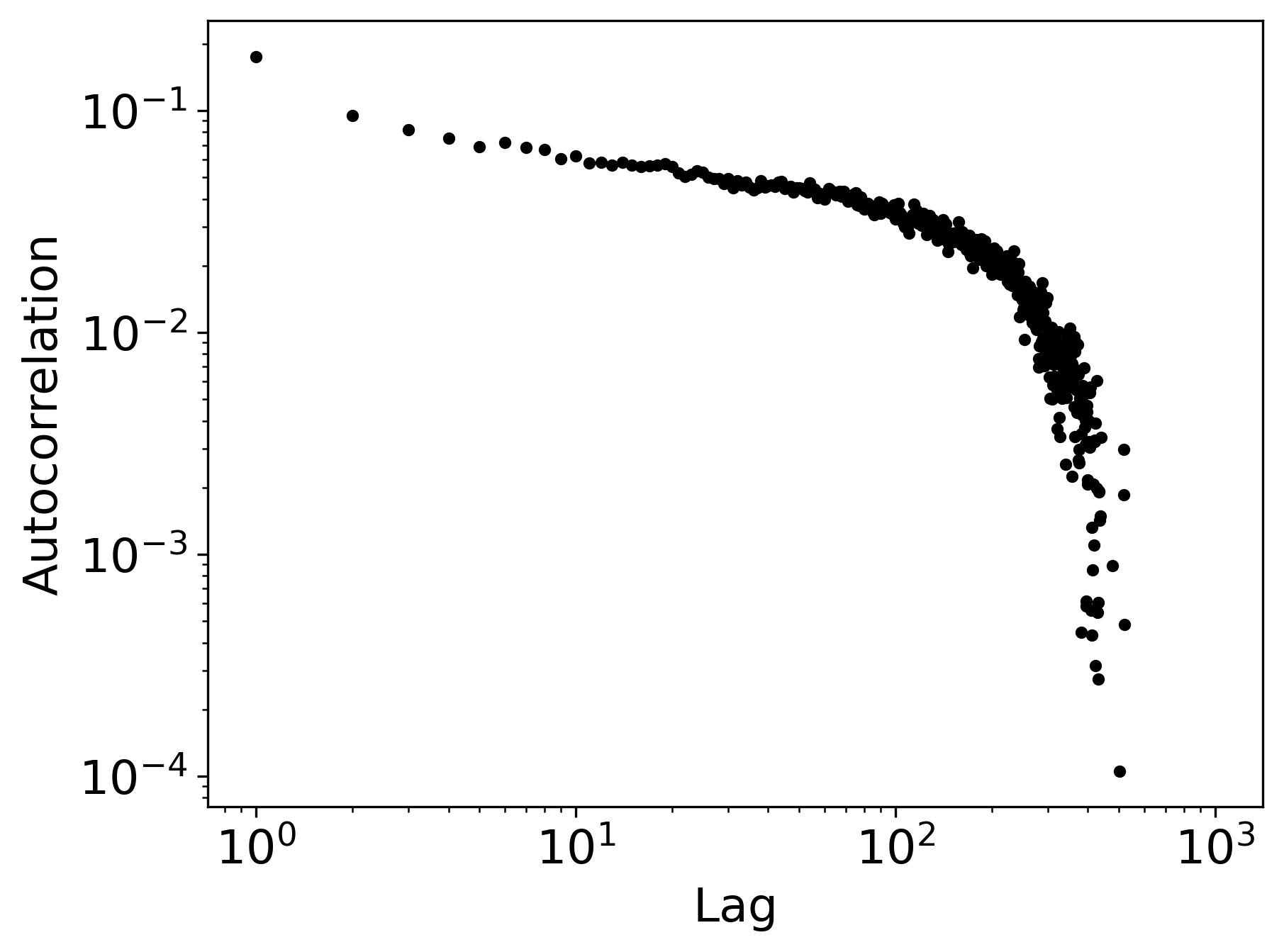}
\end{minipage}
\begin{minipage}{0.31\textwidth}
  \includegraphics[width=\linewidth]{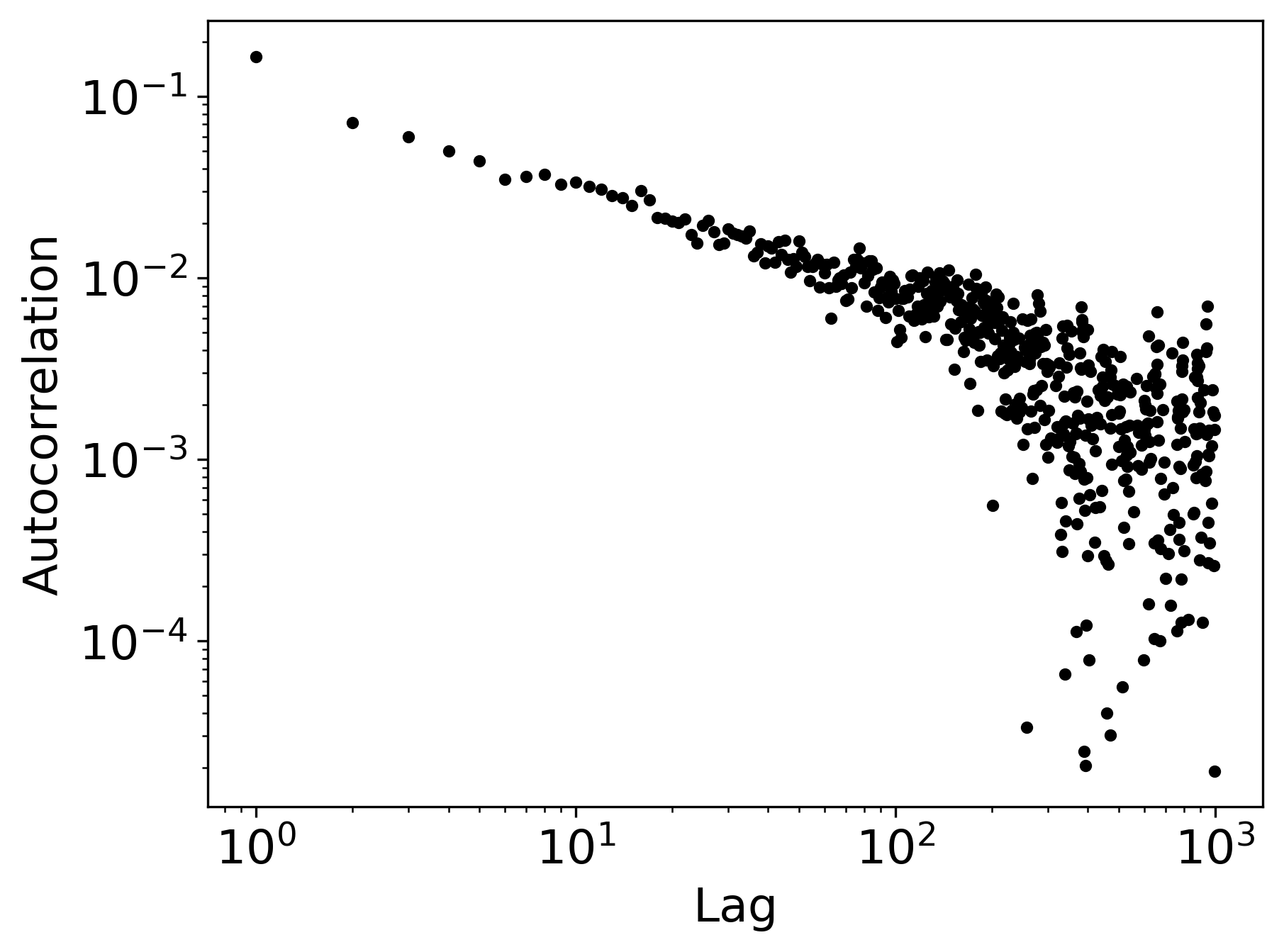}
\end{minipage}
\end{adjustbox}

\caption{
Volatility clustering across three SDE variants (rows) under different noise schedules (columns). 
}
\label{fig:volatility-clustering}
\end{figure}

Figure~\ref{fig:volatility-clustering} presents the autocorrelation of absolute log-returns for each SDE variant across three different noise schedules. In empirical financial data, such autocorrelation typically decays slowly with lag and often follows a power-law pattern, reflecting long-range dependence in volatility.

As found in the first row of Figure~\ref{fig:volatility-clustering}, VE SDE fails to capture long-range dependence, particularly under the linear schedule where the autocorrelation curve deteriorates sharply after a certain lag and displays unnatural flattening or collapse. This deficiency stems from its additive noise structure, which is independent of the state and lacks any volatility feedback mechanism. While VP SDE shares the additive noise structure of VE SDE and thus lacks a mechanism for volatility feedback, its variance-preserving formulation induces a more stable reverse process, which may help retain short-term volatility dependence to a limited extent.

Among the models considered, GBM SDE most effectively reproduces this behavior. Under both the exponential and cosine schedules, the decay pattern closely aligns with empirical observations, exhibiting a gradual and consistent decline in autocorrelation over several orders of magnitude in lag. This performance can be attributed to the multiplicative structure of the GBM model, where higher price levels lead to proportionally larger fluctuations, which naturally results in volatility clustering.

\subsection{Leverage effect}
\begin{figure}[H]
\centering

\begin{minipage}{0.05\textwidth}~\end{minipage}
\begin{minipage}{0.3\textwidth}\centering \textbf{Linear schedule} \end{minipage}
\begin{minipage}{0.3\textwidth}\centering \textbf{Exponential schedule} \end{minipage}
\begin{minipage}{0.3\textwidth}\centering \textbf{Cosine schedule} \end{minipage}

\vspace{0.5em}

\begin{adjustbox}{width=\textwidth}
\begin{minipage}{0.05\textwidth}
  \centering
  \rotatebox{90}{\textbf{VE SDE}}
\end{minipage}
\begin{minipage}{0.31\textwidth}
  \includegraphics[width=\linewidth]{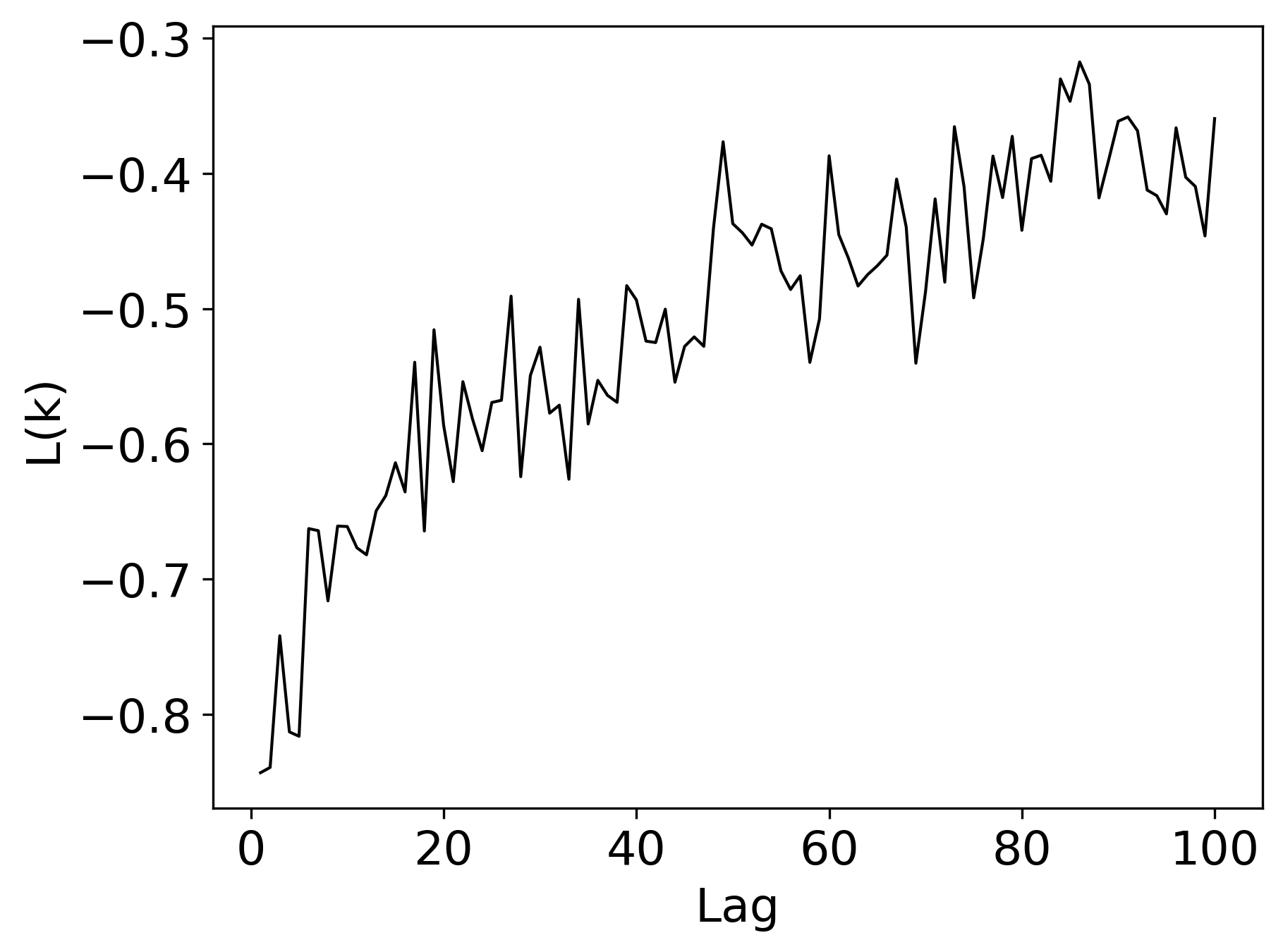}
\end{minipage}
\begin{minipage}{0.31\textwidth}
  \includegraphics[width=\linewidth]{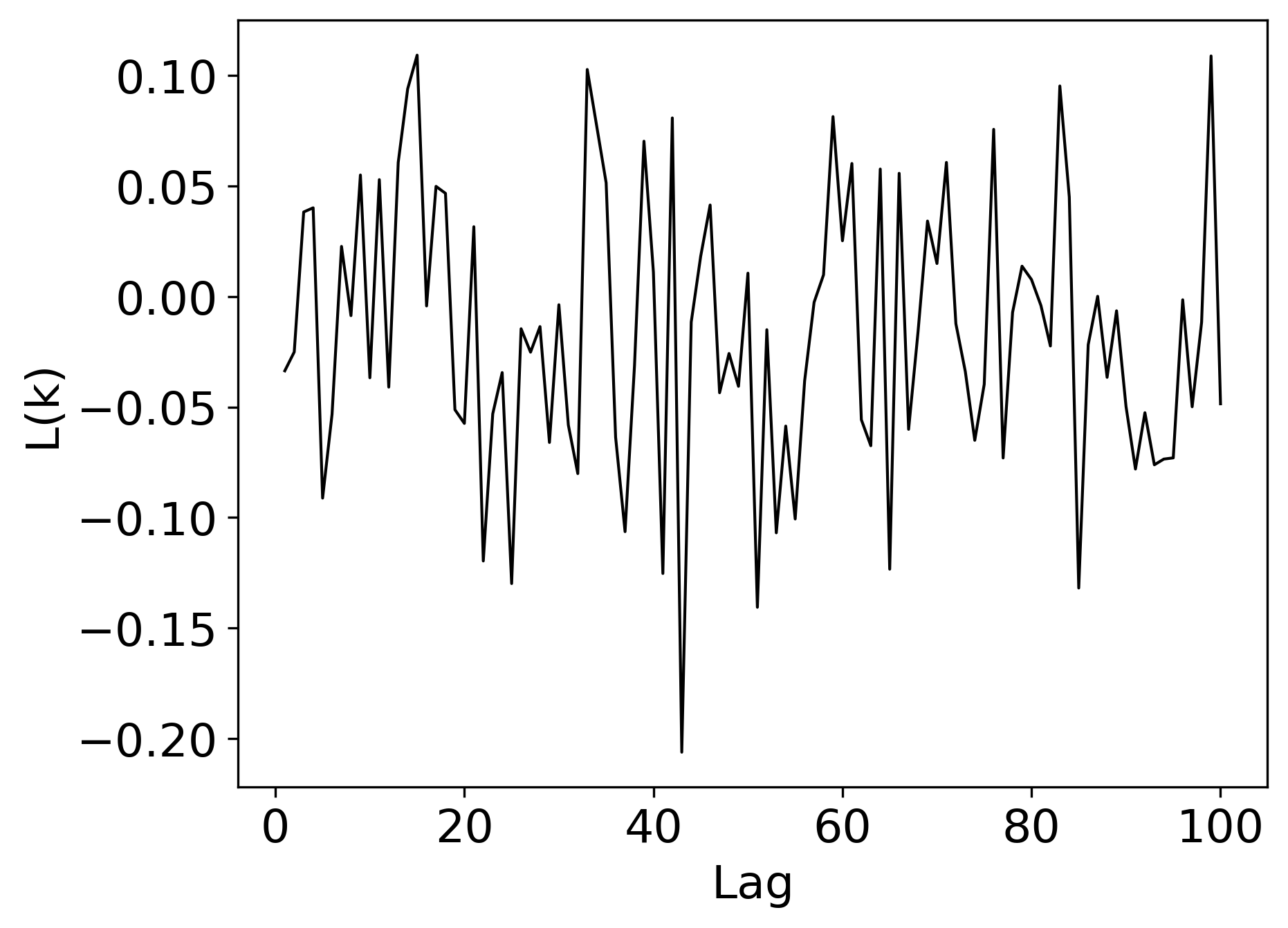}
\end{minipage}
\begin{minipage}{0.31\textwidth}
  \includegraphics[width=\linewidth]{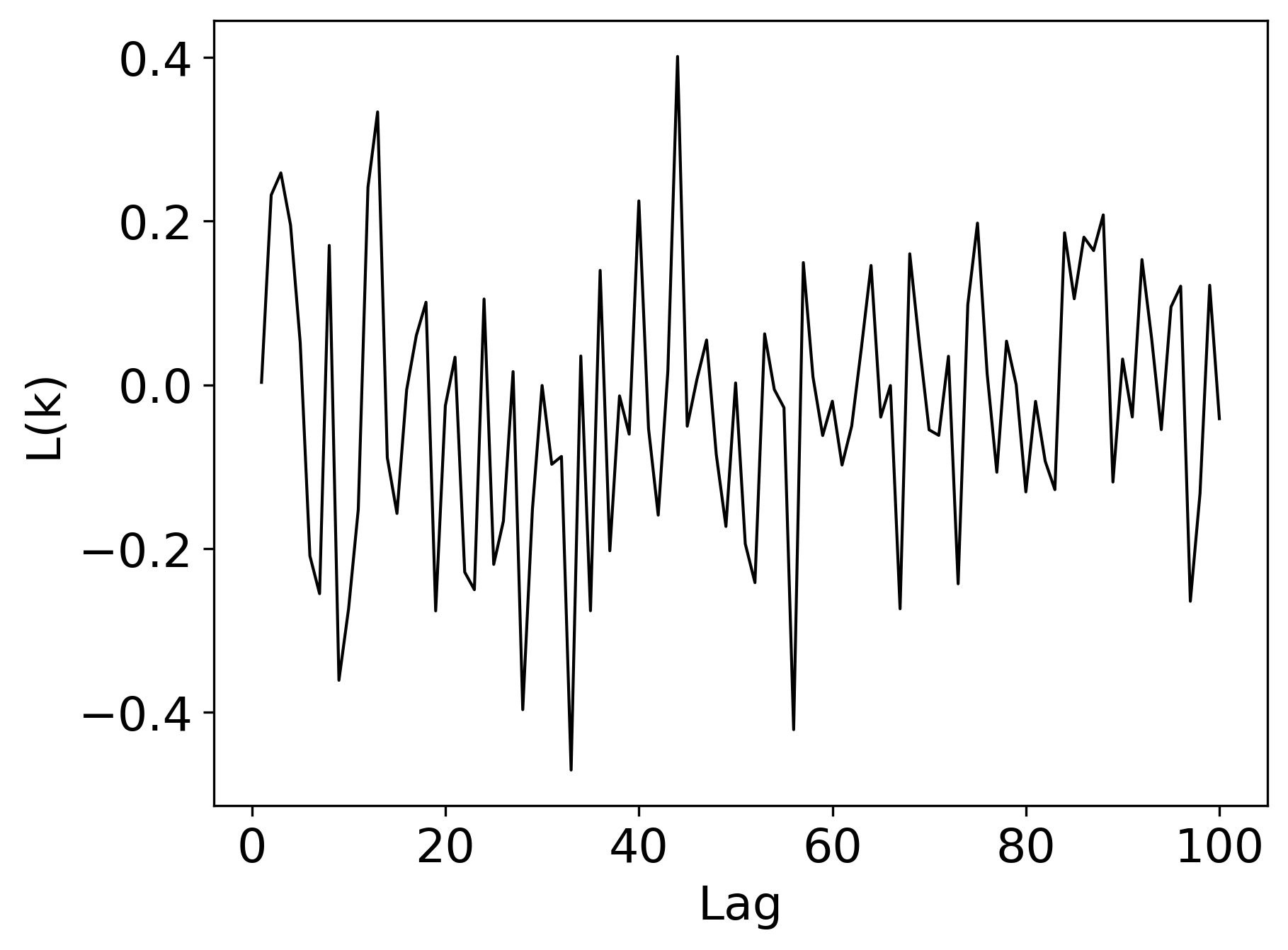}
\end{minipage}
\end{adjustbox}

\vspace{0.2em}

\begin{adjustbox}{width=\textwidth}
\begin{minipage}{0.05\textwidth}
  \centering
  \rotatebox{90}{\textbf{VP SDE}}
\end{minipage}
\begin{minipage}{0.31\textwidth}
  \includegraphics[width=\linewidth]{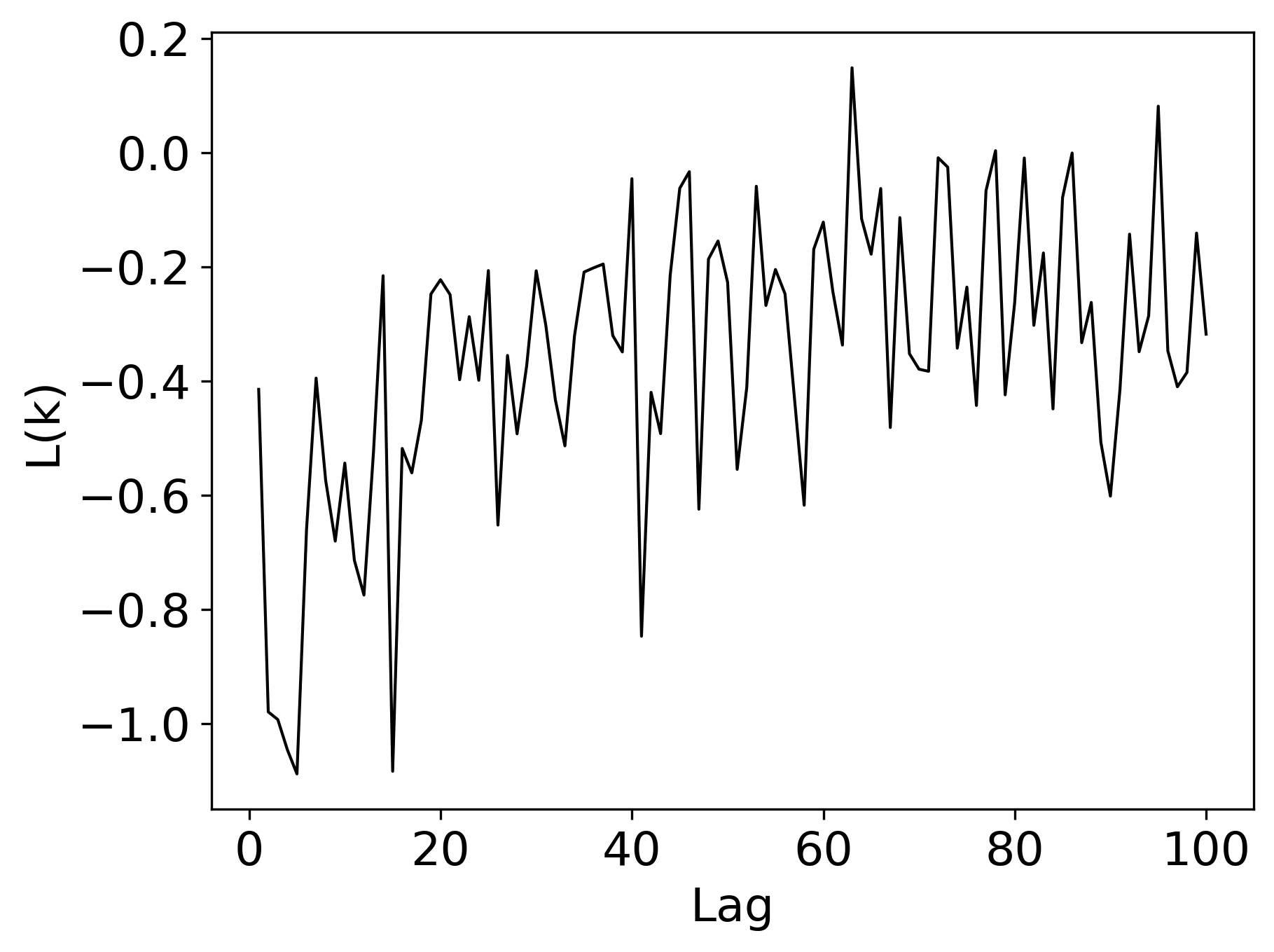}
\end{minipage}
\begin{minipage}{0.31\textwidth}
  \includegraphics[width=\linewidth]{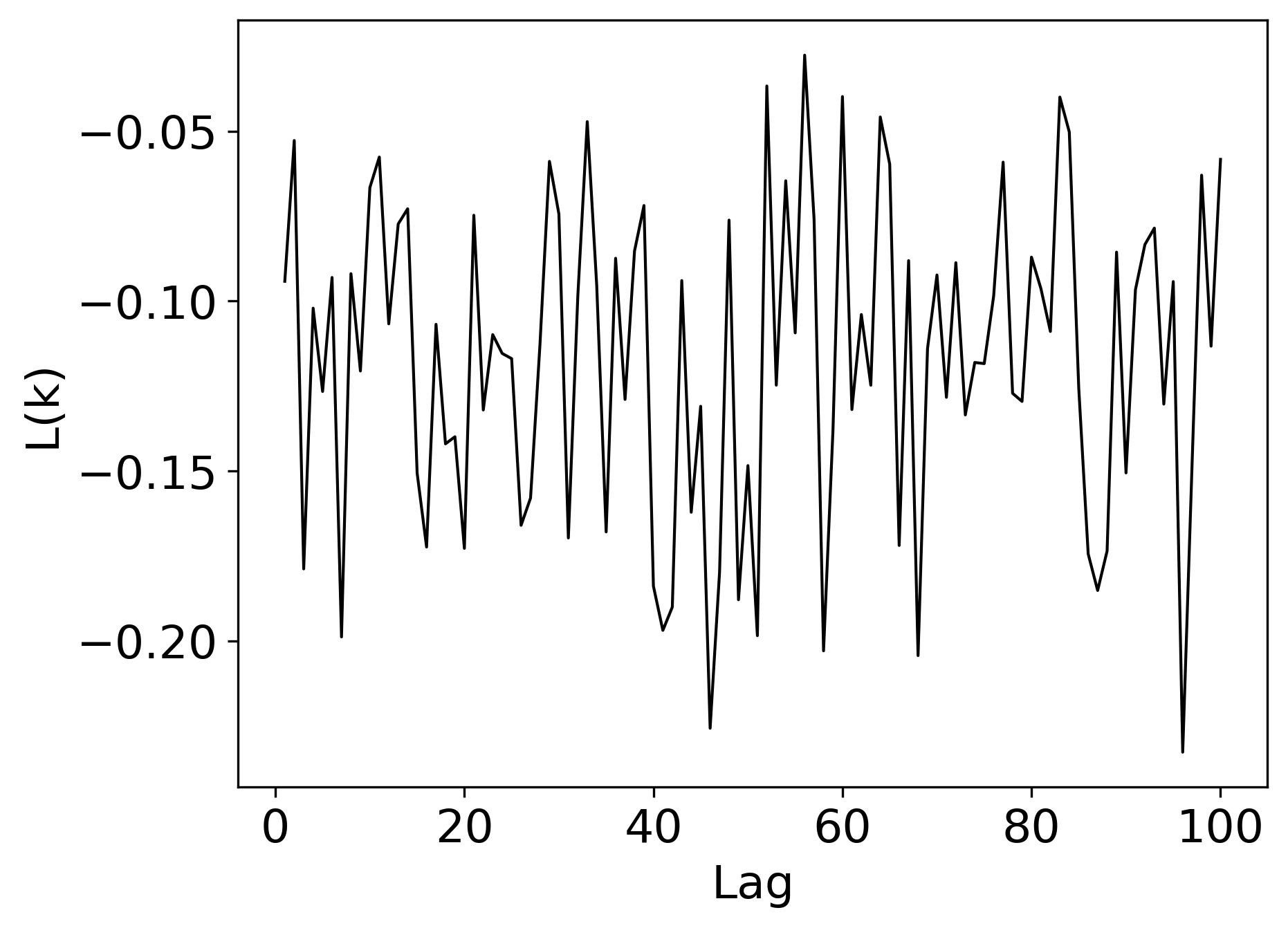}
\end{minipage}
\begin{minipage}{0.31\textwidth}
  \includegraphics[width=\linewidth]{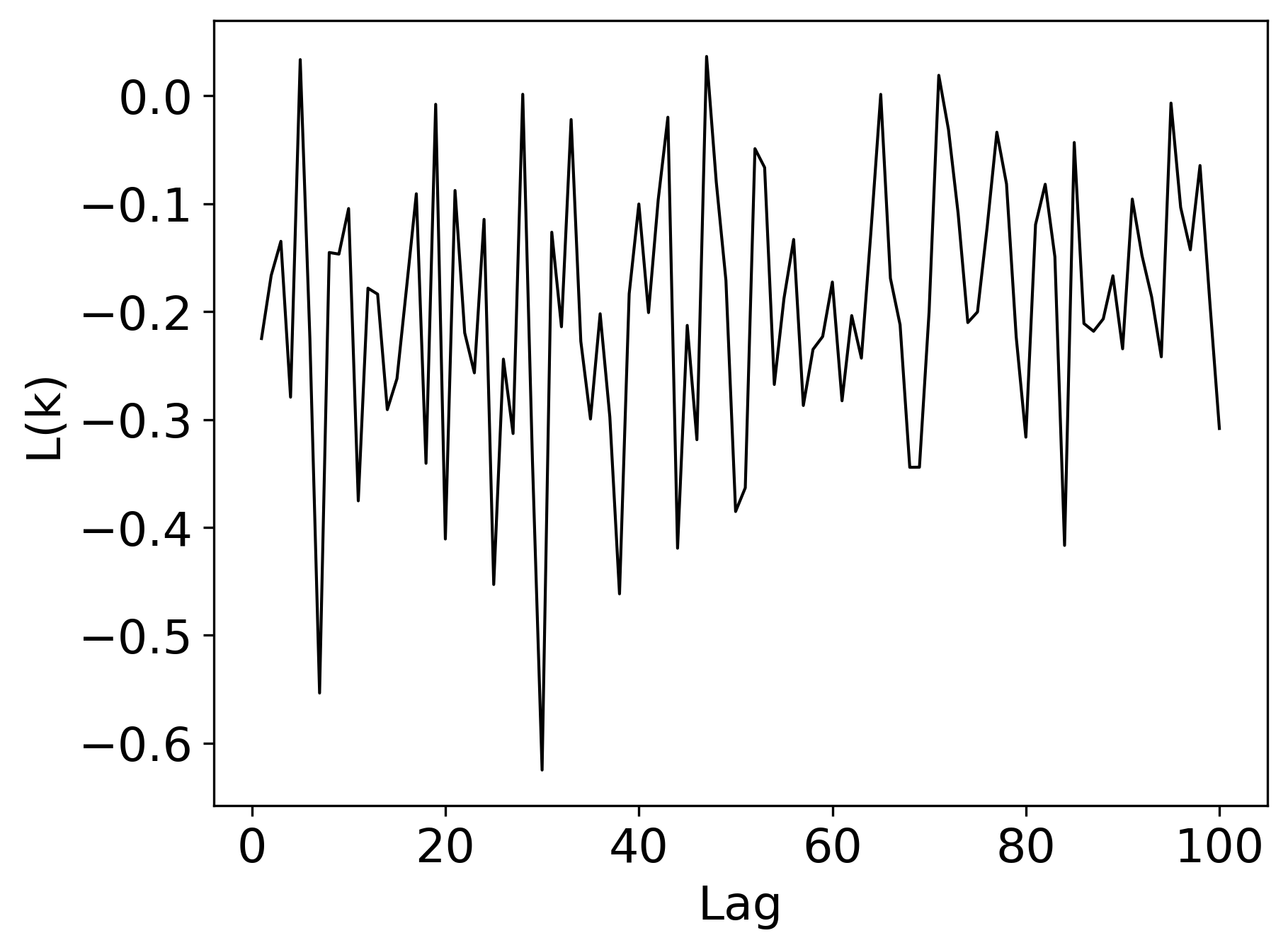}
\end{minipage}
\end{adjustbox}

\vspace{0.2em}

\begin{adjustbox}{width=\textwidth}
\begin{minipage}{0.05\textwidth}
  \centering
  \rotatebox{90}{\textbf{GBM SDE}}
\end{minipage}
\begin{minipage}{0.31\textwidth}
  \includegraphics[width=\linewidth]{figs/BS_linear_plot_64/generated_metrics_leverage_effect.png}
\end{minipage}
\begin{minipage}{0.31\textwidth}
  \includegraphics[width=\linewidth]{figs/BS_exponential_plot_64/generated_metrics_leverage_effect.png}
\end{minipage}
\begin{minipage}{0.31\textwidth}
  \includegraphics[width=\linewidth]{figs/BS_cosine_plot_64/generated_metrics_leverage_effect.png}
\end{minipage}
\end{adjustbox}

\caption{
Leverage effect across three SDE variants (rows) and noise schedules (columns).  
}
\label{fig:leverage-effect}
\end{figure}

Beyond long-range volatility dependence, another key empirical feature in financial markets is the leverage effect, characterized by negative returns being followed by increased future volatility. This effect is typically quantified by the negative correlation between returns and future squared returns. Accurately capturing this asymmetry is critical for realistic modeling of risk and volatility dynamics \cite{cont2001empirical, bouchaud2001leverage}. To assess how each SDE variant reproduces the leverage effect, we compute the lead–lag correlation between returns and future volatility, defined as the correlation between \(r_t\) and \(r_{t+k}^2\) for lags \(k = 0,\ldots,100\). 

VE SDE exhibits a mild leverage effect under the linear schedule, with a negative peak at lag $k=0$, but the correlation quickly diminishes and oscillates around zero at higher lags. The exponential and cosine schedules show weaker and noisier patterns, lacking consistent decay. On the other hand, VP SDE fails to consistently capture the leverage effect under any schedule. Under linear scheduling, the correlation fluctuates without clear structure. Exponential and cosine schedules yield slightly more centered behavior, but still lack smooth decay or persistent asymmetry. 

Similar to other stylized facts, GBM SDE most effectively reproduces leverage patterns. All three schedules exhibit a strong negative correlation at lag $k=0$, especially the cosine schedule, which drops below $-5$. Although the initial response may be amplified, the correlation stabilizes and remains moderately negative across subsequent lags, reflecting persistent asymmetry aligned with real-world leverage dynamics. This behavior arises from the fact that lower prices following negative returns result in larger future noise amplitudes under the GBM diffusion term, naturally capturing the leverage effect.


\subsection{Comparison with GAN~\cite{takahashi2019modeling} and real-world data}
\begin{figure}[H]
  \centering

  \begin{subfigure}[t]{0.32\textwidth}
    \includegraphics[width=\linewidth]{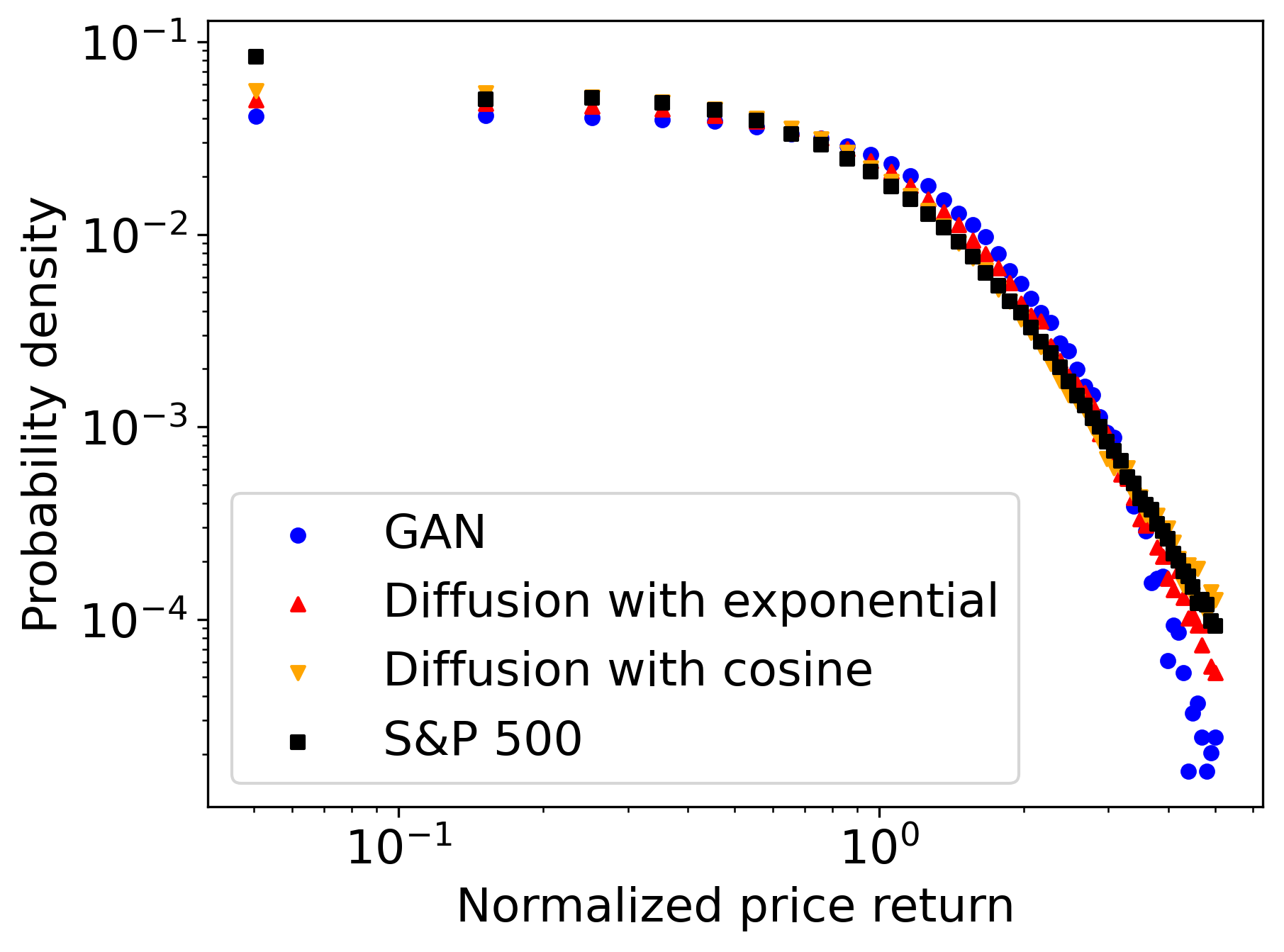}
    \caption{Heavy-tail distribution with $\alpha=4.35$}
    
  \end{subfigure}\hfill
  \begin{subfigure}[t]{0.32\textwidth}
    \includegraphics[width=\linewidth]{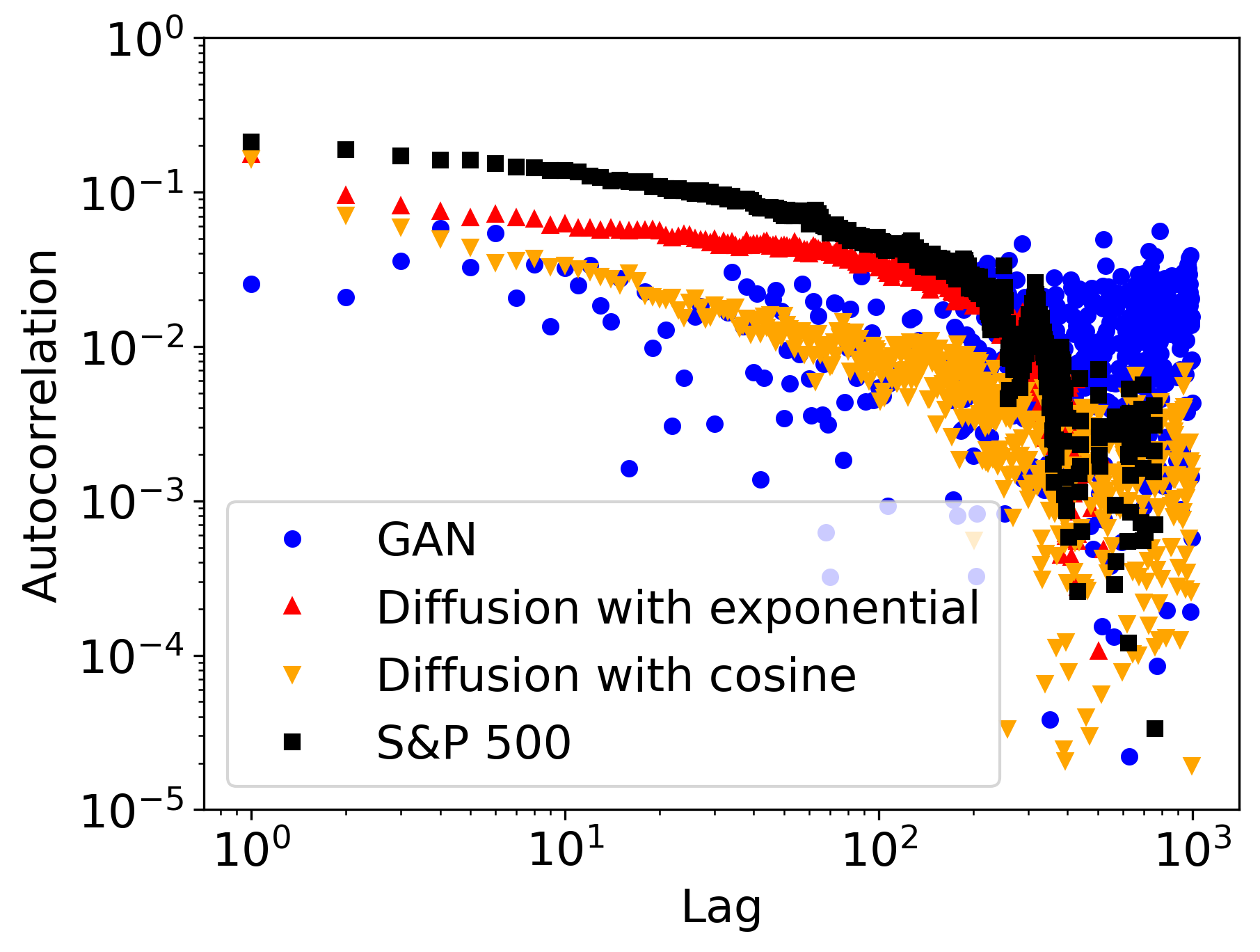}
    \caption{Volatility clustering}
    
  \end{subfigure}\hfill
  \begin{subfigure}[t]{0.32\textwidth}
    \includegraphics[width=\linewidth]{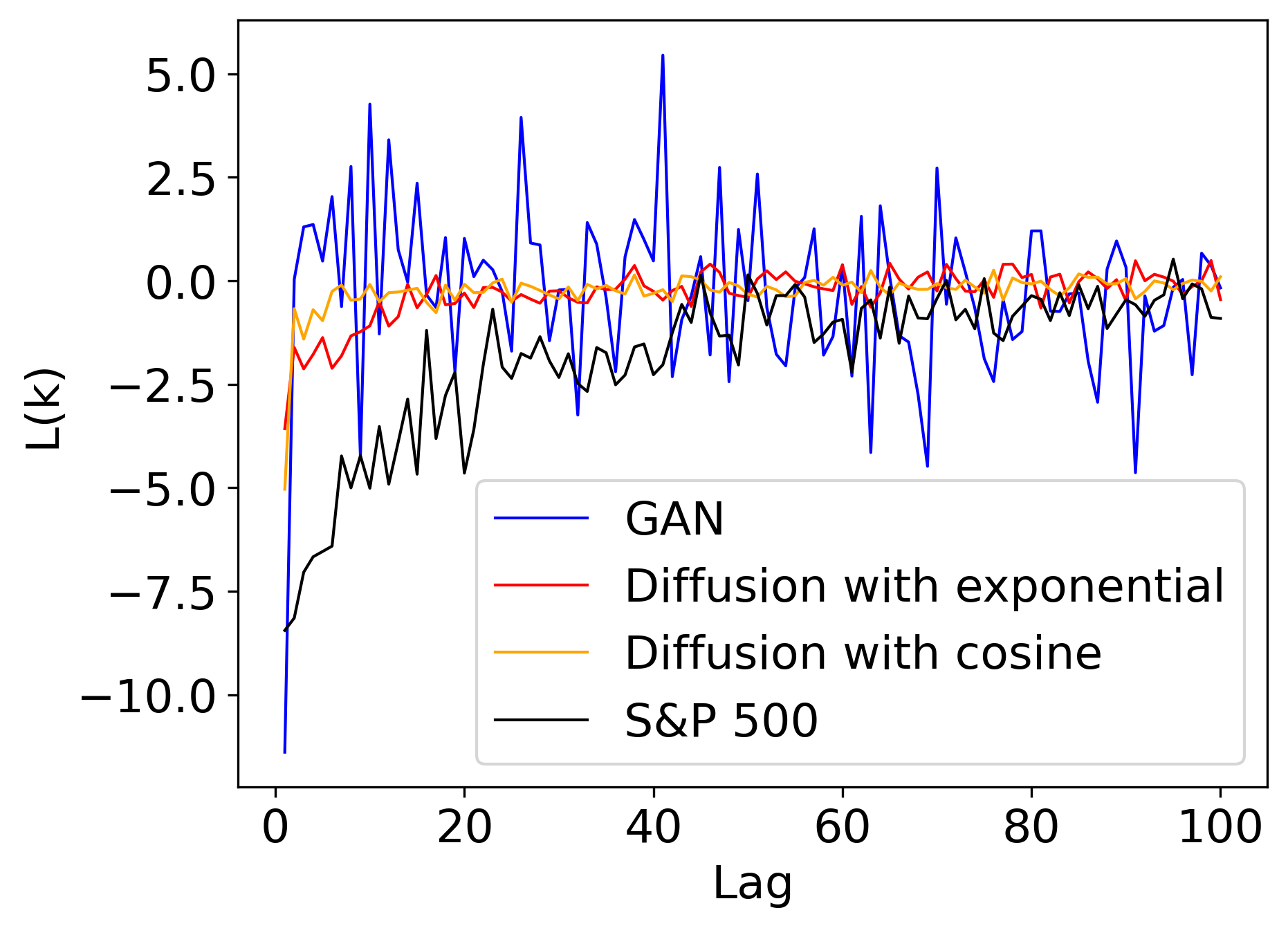}
    \caption{Leverage effect}
   
  \end{subfigure}

  \caption{
    Three statistical properties are generated by MLP generator~\cite{takahashi2019modeling}. 
}
  \label{fig:fingan}
\end{figure}
To further validate the effectiveness of our GBM-based diffusion model, we compare its performance against a GAN-based approach introduced in \cite{takahashi2019modeling}, which was specifically designed for financial time series synthesis. The comparison focuses on the three stylized facts discussed throughout the previous sections. 

Figure~\ref{fig:fingan}(a) shows the distribution of normalized returns on a log–log scale. While both models successfully reproduce the heavy-tail property with a tail exponent close to the empirical benchmark ($\alpha = 4.35$), the GBM-based diffusion model more accurately captures the extreme tails, resulting in a closer fit to real financial data. In terms of volatility clustering, Figure~\ref{fig:fingan}(b) presents the autocorrelation of squared returns. The GAN model displays short-range dependence that decays quickly and noisily. In contrast, our GBM-based model exhibits a smoother and more persistent decay pattern, closely aligned with the long-range dependence observed in empirical data. Figure~\ref{fig:fingan}(c) compares the leverage effect via lead–lag correlation between returns and future volatility. The GAN-based model exhibits erratic oscillations with no consistent structure, whereas the GBM model produces a stable, negative correlation that reflects the asymmetry characteristic of real-world leverage dynamics.

Overall, these results indicate that the proposed GBM-based diffusion model consistently outperforms the GAN-based baseline across all key evaluation metrics. By embedding financial theory into the noise structure, our model not only improves statistical fidelity but also enhances interpretability, offering a more robust foundation for realistic financial data generation.

\section{Conclusion}
In this study, we propose a novel diffusion-based generative model for financial time series by incorporating geometric Brownian motion (GBM) into the forward diffusion process. Our method is designed to better capture the stochastic and continuous-time nature of asset prices while preserving theoretical consistency with foundational financial models. Empirical results demonstrate that our diffusion model outperforms GAN-based alternatives in reproducing key stylized facts of financial markets, including heavy-tailed return distributions, volatility clustering, and the leverage effect. Furthermore, it shows strong generalization across different schedule types, validating its robustness. 

{
Forecasting time series data using the Black--Scholes model carries significant practical implications. First, most derivative pricing in financial markets is fundamentally based on the Black--Scholes framework. While extended models such as local volatility \cite{dupire1994pricing}, stochastic volatility \cite{heston1993closed}, and more recently, rough volatility \cite{gatheral2018volatility} models have gained attention, vanilla options like calls and puts are still predominantly priced using the original Black--Scholes model. Moreover, the model remains central in deriving theoretical option prices and continues to serve as a key reference in financial markets.

In addition, implied volatility—one of the most widely used volatility measures in financial markets—is derived under the Black--Scholes framework. A prominent example is the Volatility Index (VIX \footnote{\url{https://www.spglobal.com/spdji/en/vix-intro/}}), which is also calculated based on the Black--Scholes model and is extensively used as a benchmark for market volatility.

From a derivatives trading perspective, the Black--Scholes model is also essential for dynamic hedging operations. This is because core Greeks-such as delta, gamma, and theta-are defined within the Black--Scholes framework and serve as fundamental tools in managing risk and executing hedging strategies.

Building on this foundation, the present study proposes a generative framework based on a forward noising process that follows GBM. This framework is capable of generating realistic financial time series data that reflect key stylized facts observed in financial markets. The GBM-based noising process enhances both the practical relevance and scalability of the framework, making it applicable not only to derivative pricing tasks but also to the computation of risk measures such as Value-at-Risk (VaR).

In particular, this study proposes a significant paradigm shift from conventional approaches that assume a specific stochastic differential equation (SDE) process for stock prices. Instead, it focuses on modeling the underlying noise process itself. By redirecting attention from the price dynamics to the structure of the noise driving those dynamics, the proposed framework offers a novel and meaningful contribution to the literature.
}

Despite the strengths of our approach, several limitations remain. First, while the forward process injects constant-variance Gaussian noise in the log-price space, the exponential transformation into price space induces effective heteroskedasticity. This behavior is in fact desirable, as it aligns with the multiplicative nature of asset price dynamics and reflects the empirically observed state-dependent volatility. However, a more fundamental structural limitation lies in the departure from the exact log-normality of marginal price distributions. In classical GBM-based models, asset prices \( S_t \) follow a log-normal distribution due to their Markovian and continuous-time stochastic evolution. In contrast, our model jointly generates the entire log-price trajectory as a sample from a learned distribution, rather than simulating a stochastic process with time-consistent marginals. As a result, the marginal distribution of each \( S_t \) may deviate from the strict log-normal form, even though the global properties of the sequence remain consistent with stylized facts. Addressing this discrepancy could involve incorporating explicit calendar-time evolution, Markovian inductive bias, or latent stochastic volatility dynamics into future generative models.

Future work includes extending the framework to incorporate richer volatility structures, such as stochastic or rough volatility, within the generative process. Furthermore, conditioning the model on macroeconomic indicators or implied volatility surfaces may improve its applicability in derivative pricing and stress testing. Another promising direction is leveraging the model’s consistency with financial theory to develop new market simulation tools or data augmentation techniques for supervised learning tasks in quantitative finance.

\section*{Acknowledgement}
Gihun Kim and Yeoneung Kim are supported by the National Research Foundation of Korea (NRF) grant funded by the Korea government(MSIT) (RS-2023-00219980, RS-2023-00211503). The work of Sun-Yong Choi was supported by the National Research Foundation of Korea (NRF) grant funded by the Korea government (MSIT) (No. RS-2024-00454493).

\bibliographystyle{IEEEtran}

\bibliography{ref}

\begin{thebibliography}{10}
\providecommand{\url}[1]{#1}
\csname url@samestyle\endcsname
\providecommand{\newblock}{\relax}
\providecommand{\bibinfo}[2]{#2}
\providecommand{\BIBentrySTDinterwordspacing}{\spaceskip=0pt\relax}
\providecommand{\BIBentryALTinterwordstretchfactor}{4}
\providecommand{\BIBentryALTinterwordspacing}{\spaceskip=\fontdimen2\font plus
\BIBentryALTinterwordstretchfactor\fontdimen3\font minus \fontdimen4\font\relax}
\providecommand{\BIBforeignlanguage}[2]{{%
\expandafter\ifx\csname l@#1\endcsname\relax
\typeout{** WARNING: IEEEtran.bst: No hyphenation pattern has been}%
\typeout{** loaded for the language `#1'. Using the pattern for}%
\typeout{** the default language instead.}%
\else
\language=\csname l@#1\endcsname
\fi
#2}}
\providecommand{\BIBdecl}{\relax}
\BIBdecl

\bibitem{Goo14}
I.~Goodfellow, J.~Pouget-Abadie, M.~Mirza, B.~Xu, D.~Warde-Farley, S.~Ozair, A.~Courville, and Y.~Bengio, ``Generative adversarial nets,'' \emph{Advances in {N}eural {I}nformation {P}rocessing Systems {(NeurIPS)}}, pp. 2672--2680, 2014.

\bibitem{kingma2013auto}
D.~P. Kingma, M.~Welling \emph{et~al.}, ``Auto-encoding variational bayes,'' 2013.

\bibitem{ho2020denoising}
J.~Ho, A.~Jain, and P.~Abbeel, ``Denoising diffusion probabilistic models,'' \emph{Advances in {N}eural {I}nformation {P}rocessing Systems {(NeurIPS)}}, vol.~33, pp. 6840--6851, 2020.

\bibitem{song2020score}
Y.~Song, J.~Sohl-Dickstein, D.~P. Kingma, A.~Kumar, S.~Ermon, and B.~Poole, ``Score-based generative modeling through stochastic differential equations,'' \emph{arXiv preprint arXiv:2011.13456}, 2020.

\bibitem{black1973pricing}
F.~Black and M.~Scholes, ``The pricing of options and corporate liabilities,'' \emph{Journal of {P}olitical {E}conomy}, vol.~81, no.~3, pp. 637--654, 1973.

\bibitem{tashiro2021csdi}
Y.~Tashiro, J.~Song, Y.~Song, and S.~Ermon, ``Csdi: {C}onditional score-based diffusion models for probabilistic time series imputation,'' \emph{Advances in {N}eural {I}nformation {P}rocessing Systems {(NeurIPS)}}, vol.~34, pp. 24\,804--24\,816, 2021.

\bibitem{takahashi2019modeling}
S.~Takahashi, Y.~Chen, and K.~Tanaka-Ishii, ``Modeling financial time-series with generative adversarial networks,'' \emph{Physica A: Statistical Mechanics and its Applications}, vol. 527, p. 121261, 2019.

\bibitem{tovar2020deep}
W.~Tovar, ``Deep learning based on generative adversarial and convolutional neural networks for financial time series predictions,'' \emph{arXiv preprint arXiv:2008.08041}, 2020.

\bibitem{rizzato2023generative}
M.~Rizzato, J.~Wallart, C.~Geissler, N.~Morizet, and N.~Boumlaik, ``Generative adversarial networks applied to synthetic financial scenarios generation,'' \emph{Physica A: Statistical Mechanics and its Applications}, vol. 623, p. 128899, 2023.

\bibitem{li2024enhancing}
S.~Li and S.~Xu, ``Enhancing stock price prediction using gans and transformer-based attention mechanisms,'' \emph{Empirical {E}conomics}, pp. 1--31, 2024.

\bibitem{vuletic2024fin}
M.~Vuleti{\'c}, F.~Prenzel, and M.~Cucuringu, ``Fin-gan: Forecasting and classifying financial time series via generative adversarial networks,'' \emph{Quantitative {F}inance}, vol.~24, no.~2, pp. 175--199, 2024.

\bibitem{richemond2022categorical}
P.~H. Richemond, S.~Dieleman, and A.~Doucet, ``Categorical sdes with simplex diffusion,'' \emph{arXiv preprint arXiv:2210.14784}, 2022.

\bibitem{ouyang2023missdiff}
Y.~Ouyang, L.~Xie, C.~Li, and G.~Cheng, ``Missdiff: {T}raining diffusion models on tabular data with missing values,'' \emph{arXiv preprint arXiv:2307.00467}, 2023.

\bibitem{anderson1982reverse}
B.~D. Anderson, ``Reverse-time diffusion equation models,'' \emph{Stochastic {P}rocesses and their Applications}, vol.~12, no.~3, pp. 313--326, 1982.

\bibitem{harmantzis2006empirical}
F.~C. Harmantzis, L.~Miao, and Y.~Chien, ``Empirical study of value-at-risk and expected shortfall models with heavy tails,'' \emph{The journal of risk finance}, vol.~7, no.~2, pp. 117--135, 2006.

\bibitem{jacquier2002bayesian}
E.~Jacquier, N.~G. Polson, and P.~E. Rossi, ``Bayesian analysis of stochastic volatility models,'' \emph{Journal of Business \& Economic Statistics}, vol.~20, no.~1, pp. 69--87, 2002.

\bibitem{dennis2006stock}
P.~Dennis, S.~Mayhew, and C.~Stivers, ``Stock returns, implied volatility innovations, and the asymmetric volatility phenomenon,'' \emph{Journal of Financial and Quantitative Analysis}, vol.~41, no.~2, pp. 381--406, 2006.

\bibitem{kong2020diffwave}
Z.~Kong, W.~Ping, J.~Huang, K.~Zhao, and B.~Catanzaro, ``Diffwave: {A} versatile diffusion model for audio synthesis,'' \emph{arXiv preprint arXiv:2009.09761}, 2020.

\bibitem{cont2001empirical}
R.~Cont, ``Empirical properties of asset returns: stylized facts and statistical issues,'' \emph{Quantitative finance}, vol.~1, no.~2, p. 223, 2001.

\bibitem{bouchaud2001leverage}
J.-P. Bouchaud, A.~Matacz, and M.~Potters, ``Leverage effect in financial markets: The retarded volatility model,'' \emph{Physical {R}eview {L}etters}, vol.~87, no.~22, p. 228701, 2001.

\bibitem{dupire1994pricing}
B.~Dupire \emph{et~al.}, ``Pricing with a smile,'' \emph{Risk}, vol.~7, no.~1, pp. 18--20, 1994.

\bibitem{heston1993closed}
S.~L. Heston, ``A closed-form solution for options with stochastic volatility with applications to bond and currency options,'' \emph{The {R}eview of {F}inancial {S}tudies}, vol.~6, no.~2, pp. 327--343, 1993.

\bibitem{gatheral2018volatility}
J.~Gatheral, T.~Jaisson, and M.~Rosenbaum, ``Volatility is rough,'' \emph{Quantitative {F}inance}, vol.~18, no.~6, pp. 933--949, 2018.

\end{thebibliography}

\end{document}